\documentclass{article}

\usepackage{arxiv}

\usepackage[utf8]{inputenc} % allow utf-8 input
\usepackage[T1]{fontenc}    % use 8-bit T1 fonts
\usepackage{hyperref}       % hyperlinks
\usepackage{url}            % simple URL typesetting
\usepackage{booktabs}       % professional-quality tables
\usepackage{amsfonts}       % blackboard math symbols
\usepackage{nicefrac}       % compact symbols for 1/2, etc.
\usepackage{microtype}      % microtypography
\usepackage{lipsum}
\usepackage{graphicx}
\graphicspath{ {./images/} }

\usepackage{amssymb,amsmath,amsthm,bm}
\usepackage{geometry}
\usepackage{algorithm}
\usepackage{algpseudocode}
\usepackage{tabularx}
\usepackage{graphicx}
\usepackage[retainorgcmds]{IEEEtrantools}
\graphicspath{{figures-pdf/}}
\usepackage[normalem]{ulem}
\usepackage{url}
\usepackage{lineno}
\usepackage{makecell}

\usepackage{multirow} % for cmd 'multirow', 'multicolumn'

\usepackage{bm}

\usepackage{color,overpic}
\usepackage{epstopdf}
\usepackage{float}

\usepackage{CJKutf8}
\usepackage{diagbox}
\usepackage{stfloats}
\usepackage{bbding}

\newlength\imagewidth
\newlength\figwidth
\newlength\sfigwidth
\newlength\vfigskip
\usepackage{multirow}

\usepackage{color}
\definecolor{dgreen}{rgb}{0,.6,0}
\usepackage{lineno}
\usepackage{float}
\usepackage{hyperref}
\usepackage{gensymb}
\usepackage[english]{babel}

\usepackage[font=small,skip=0pt]{caption}

\title{EEG-Driven 3D Object Reconstruction with Style Consistency\\ and Diffusion Prior}

\author{
 Xin Xiang \\
  School of Computer Science and Technology\\
  Hangzhou Dianzi University\\
  Hangzhou \\
  \texttt{} \\
   \And
 Wenhui Zhou \\
  School of Computer Science and Technology\\
  Hangzhou Dianzi University\\
  Hangzhou \\
  \texttt{} \\
  \And
 Guojun Dai \\
  School of Computer Science and Technology\\
  Hangzhou Dianzi University\\
  Hangzhou \\
  \texttt{} \\
}

\begin{document}
\maketitle
\begin{abstract}
Electroencephalography (EEG) -based visual perception reconstruction has become an important area of research. Neuroscientific studies indicate that humans can decode imagined 3D objects by perceiving or imagining various visual information, such as color, shape, and rotation. Existing EEG-based visual decoding methods typically focus only on the reconstruction of 2D visual stimulus images and face various challenges in generation quality, including inconsistencies in texture, shape, and color between the visual stimuli and the reconstructed images. This paper proposes an EEG-based 3D object reconstruction method with style consistency and diffusion priors. The method consists of an EEG-driven multi-task joint learning stage and an EEG-to-3D diffusion stage. The first stage uses a neural EEG encoder based on regional semantic learning, employing a multi-task joint learning scheme that includes a masked EEG signal recovery task and an EEG based visual classification task. The second stage introduces a latent diffusion model (LDM) fine-tuning strategy with style-conditioned constraints and a neural radiance field (NeRF) optimization strategy. This strategy explicitly embeds semantic- and location-aware latent EEG codes and combines them with visual stimulus maps to fine-tune the LDM. The fine-tuned LDM serves as a diffusion prior, which, combined with the style loss of visual stimuli, is used to optimize NeRF for generating 3D objects. Finally, through experimental validation, we demonstrate that this method can effectively use EEG data to reconstruct 3D objects with style consistency.
\end{abstract}

\section{Introduction}
A noninvasive brain-computer interface (BCI) system typically controls electronic devices through voluntary modulation of EEG signals. However, most current studies investigating the relationship between brain activity and 3D object visual imagery tasks primarily rely on functional magnetic resonance imaging (fMRI). For example, Mind-3D ~\cite{gao2023mind} successfully decoded 3D visual information from the brain using fMRI signals, demonstrating the feasibility of this challenging task. Therefore, existing fMRI studies provide theoretical foundations and research support for exploring different neuroimaging techniques and understanding the neural modulation in 3D object imagery tasks, especially in the context of EEG-based decoding of 3D visual information from the brain.
\begin{figure*}
    \centering
    \includegraphics[width=0.9\linewidth]{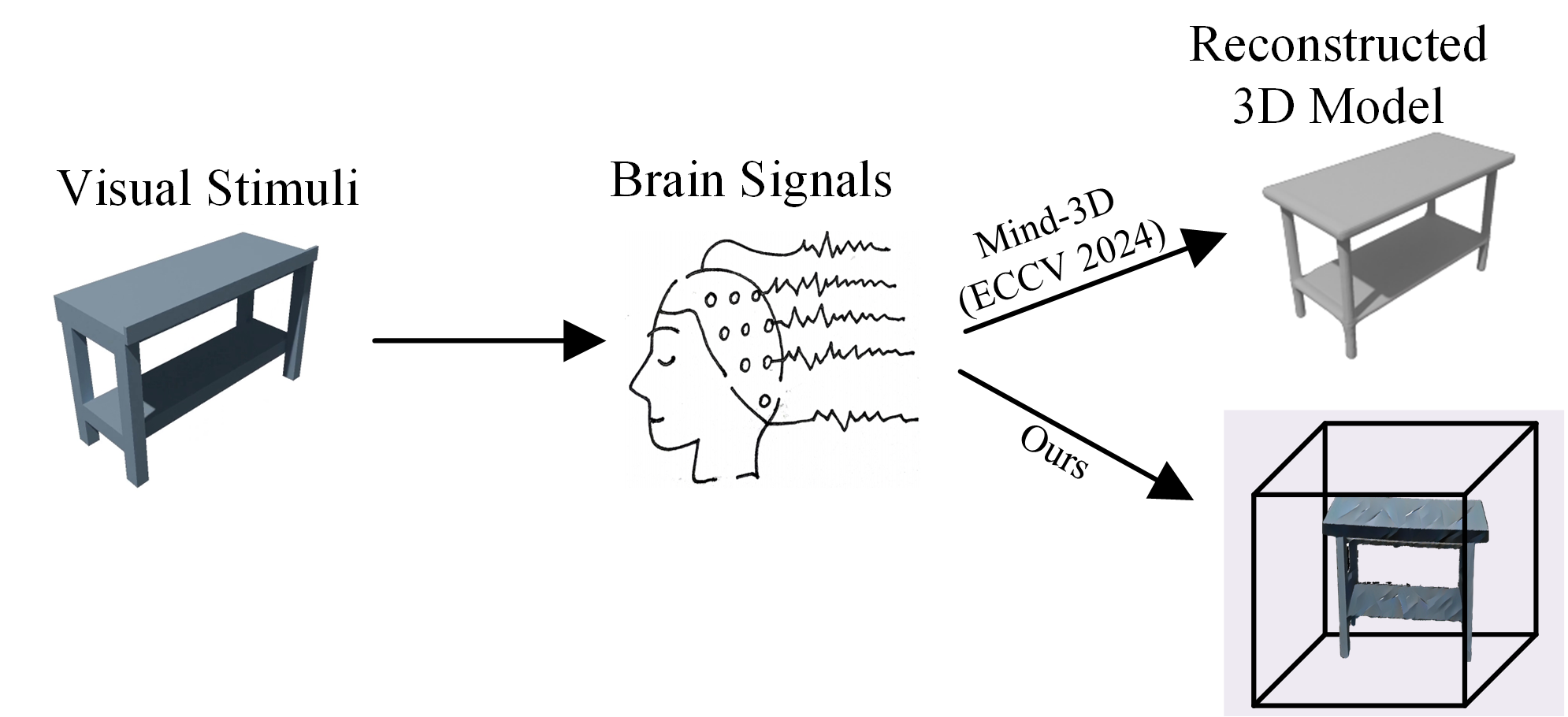}
    \caption{Our model reconstructs 3D objects using fMRI signals.}
    \label{introduction}
\end{figure*}
\label{sec:intro}
A large body of neuroscientific experimental research ~\cite{spampinato2017deep, hegde2008time, fabre2003visual, vanrullen2003competition, abrams1989speed, rayner1978eye, chen1982topological} has demonstrated that the brain can rapidly perceive 2D images and 3D objects in extremely short periods of time. One of the primary methods for studying human visual perception is using deep neural networks to reconstruct visual content that evokes subjective responses in stimulus experiments. Many studies ~\cite{Chen_2023_CVPR, du2023decoding, lu2023minddiffuser, takagi2023high} have attempted to reconstruct visual information based on functional magnetic resonance imaging (fMRI). However, the collection of fMRI data requires expensive equipment, limiting its widespread use in practical applications. In contrast, EEG is a more cost-effective technique for capturing brain activity and is easier to collect. EEG data are typically recorded as a series of time-series electrophysiological signals by placing electrodes on the scalp. During this process, subjects are presented with stimulus images, while brain signals are simultaneously recorded.

Recently, several studies have explored human visual perception learning based on EEG signals. For example, ~\cite{kavasidis2017brain2image} used traditional generative models to convert EEG signals into images, while ~\cite{BaiECCV24} and ~\cite{ye2024self} utilized latent diffusion models (LDM) to extract latent features for reconstructing visual stimulus images corresponding to EEG signals, achieving better alignment between EEG signals and 2D images. However, these attempts still have limitations in terms of pixel-level and semantic fidelity, and no research to date has utilized EEG signals to reconstruct color-consistent 3D objects. One reason for this is the difficulty in effectively capturing semantic information, and another is the relative complexity of the learning process in generative models. As a result, the reconstructed images often lack accurate perceptual information and struggle to achieve precise prediction and control of structural details.

Building upon the aforementioned neuroscientific theories and technological advancements, it can be inferred that reconstructing high-quality visual information from brain activity is feasible. In this study, we propose a method for reconstructing color-consistent 3D objects from EEG signals. To address the challenges of this task, we jointly train EEG signals with Ground Truth (GT) images. Specifically, we first train an implicit neural EEG encoder with the capability of perceiving 3D objects, allowing it to capture regional semantic features. Then, based on the latent EEG codes obtained in the first stage, we integrate a diffusion model, neural style loss, and NeRF to implicitly decode the 3D objects. Finally, through experimental validation, we demonstrate that our method is capable of reconstructing color-consistent 3D objects using EEG signals.

In order for the implicit neural EEG encoder to capture regional semantic features, in the first stage, we employ joint training for reconstruction and classification using EEG signals, enabling the EEG encoder to learn the regional semantic features of EEG. The reconstruction task aims to reconstruct the EEG signals, learning the temporal information associated with specific regions, while the semantic classification task is used to classify the semantic features of the EEG signals, training the encoder to recognize semantic characteristics. Through the joint learning of these two tasks, the proposed method can effectively capture regional semantic features.

In order for the implicit neural decoder to decode 3D objects, in the second stage, this paper integrates latent EEG codes, a diffusion model, and NeRF to decode 3D objects. The latent EEG codes guide the diffusion model to generate novel 2D views, and a style loss is used to transfer the colors from GT, ensuring that the colors of the novel views remain as consistent with the GT as possible. Subsequently, the novel 2D views from different perspectives are used to optimize NeRF. Unlike previous text-to-3D methods ~\cite{poole2022dreamfusion, tang2023make}, the proposed approach focuses on reconstructing color-consistent 3D objects based on latent EEG encodings. 

% \textcolor{red}{To enable NeRF to more effectively learn 3D information, a powerful depth estimation model is used to obtain depth information from different views, assisting the NeRF model in better understanding the 3D structure of the scene during the optimization process.}

The main contributions of this paper are as follows:
\begin{itemize}
    \item We propose an EEG-based 3D object reconstruction framework with style-consistent semantic region awareness for reconstructing 3D objects that are consistent with the style of visual stimuli. The framework consists of an EEG multi-task joint learning stage and a style-semantic region-aware LDM fine-tuning and NeRF optimization stage. The former focuses on learning semantic- and location-aware latent EEG codes from EEG signals, while the latter uses these learned latent EEG codes as conditions to fine-tune LDM. The fine-tuned LDM serves as a diffusion prior, which, in combination with a style loss, optimizes NeRF and ultimately reconstructs style-consistent 3D objects.
    \item We design a neural EEG encoder based on regional semantic learning, employing a multi-task joint learning scheme that includes a masked EEG signal recovery task and an EEG-based visual classification task. According to the visual attention shifting mechanism, the fixation regions of the human eye change over time. Thus, the recovery task helps the EEG encoder learn the spatial location information of objects by reconstructing missing EEG data, while the classification task aids in learning the semantic information of object regions.
   \item This paper proposes a fine-tuning strategy for LDM with style-conditioned constraints and NeRF optimization. The strategy explicitly embeds semantic- and location-aware latent EEG codes and incorporates visual stimulus maps to fine-tune the LDM. The fine-tuned LDM serves as a diffusion prior, which, combined with the style loss of visual stimuli, is used to optimize NeRF. Finally, EEG data is employed to reconstruct 3D objects with color consistency.
\end{itemize}

\section{Related Work} \label{section2}
\paragraph{\textbf{EEG-Based Image Synthesis}}Over the past few years, text-to-image generation models have developed rapidly. For example, diffusion models have made significant progress in this field, as they are capable of extracting complex latent semantic features from text descriptions and generating high-quality object and scene images ~\cite{ruiz2023dreambooth, saharia2022photorealistic, sanghi2022clip, schwarz2022voxgraf, BaiECCV24}. In this paper, we fine-tune the diffusion model using EEG-ImageNet ~\cite{spampinato2017deep} and Things-EEG2 ~\cite{gifford2022large} dataset to achieve the task of generating 2D images from EEG signals. For instance, DreamDiffusion ~\cite{BaiECCV24} fine-tunes the diffusion model through global semantic alignment, but there still remains a certain gap between the generated images and the actual images. To enhance the realism of the generated images, the proposed method combines EEG signal reconstruction and classification to help the model capture regional semantic features, enabling the diffusion model to better learn the regional mapping relationship between EEG signals and images.

% \paragraph{\textbf{Single-View 3D Objects Modeling}}Traditional 3D object reconstruction methods typically require images from different viewpoints to reconstruct the three-dimensional structure. However, in recent years, methods leveraging deep learning, computer vision, and NeRF have emerged, enabling the generation of 3D objects from a single image. Research on 3D generation models involves various types of 3D representation methods, including 3D voxel grids ~\cite{smith2017improved, sohl2015deep, song2019generative, takikawa2021neural} and point clouds ~\cite{wu2016learning, yang2019pointflow, vahdat2022lion}, among others. Benefiting from the success of neural volumetric rendering techniques such as NeRF, approaches that utilize 2D pretrained models to optimize 3D models have gradually emerged, reducing the dependence on multi-view data. For example, methods like DreamFusion ~\cite{poole2022dreamfusion}, Magic3D ~\cite{lin2023magic3d}, and Make-It-3D ~\cite{tang2023make} reconstruct 3D objects from a single viewpoint, combining zero-shot learning methods guided by CLIP ~\cite{radford2021learning} to constrain and optimize NeRF.

\paragraph{\textbf{EEG-to-3D Object Generation}}In recent years, significant progress has been made in text-to-3D object generation models ~\cite{lin2023magic3d, poole2022dreamfusion, tang2023make} , but EEG-to-3D object generation has not yet been fully explored. Inspired by text-to-3D methods such as ~\cite{poole2022dreamfusion} and ~\cite{tang2023make}, this paper attempts to utilize latent EEG codes to guide NeRF in reconstructing 3D objects through the shape priors of a diffusion model. To further investigate the brain's ability to perceive color visual information from EEG, and drawing upon content and style transfer methods such as ~\cite{li2018closed} and ~\cite{chiu2022pca}, we employ content and style losses to ensure that the reconstructed 3D objects maintain style consistency with ground truth (GT) images. Using this approach, we have, for the first time, achieved the reconstruction of style-consistent 3D objects from EEG signals, indirectly validating the neuroscientific theory that the human brain can perceive various types of visual information, such as color, shape, and texture, when observing objects.

% \paragraph{\textbf{Fine-Tuning}}In the era of large models, leveraging the weights of pretrained models has become increasingly important. Since these models have already been trained on vast datasets, we can utilize their rich representational capabilities to enhance the performance of our models on new tasks. For example, through contrastive learning other pretext tasks, we can acquire contextual knowledge for downstream tasks. Taking Stable Diffusion (SD) as an example, it is trained on the LAION-5B dataset, which consists of 5 billion image-text pairs. Since our dataset is based on EEG-ImageNet ~\cite{spampinato2017deep}, we opted to use SD's parameters as the initial parameters, followed by further training on our dataset to better adapt the SD model to the new task.

\renewcommand{\dblfloatpagefraction}{.9}
\begin{figure*}[ht]
    \centering % 设置图片居中对齐
    \includegraphics[width=0.85\textwidth]{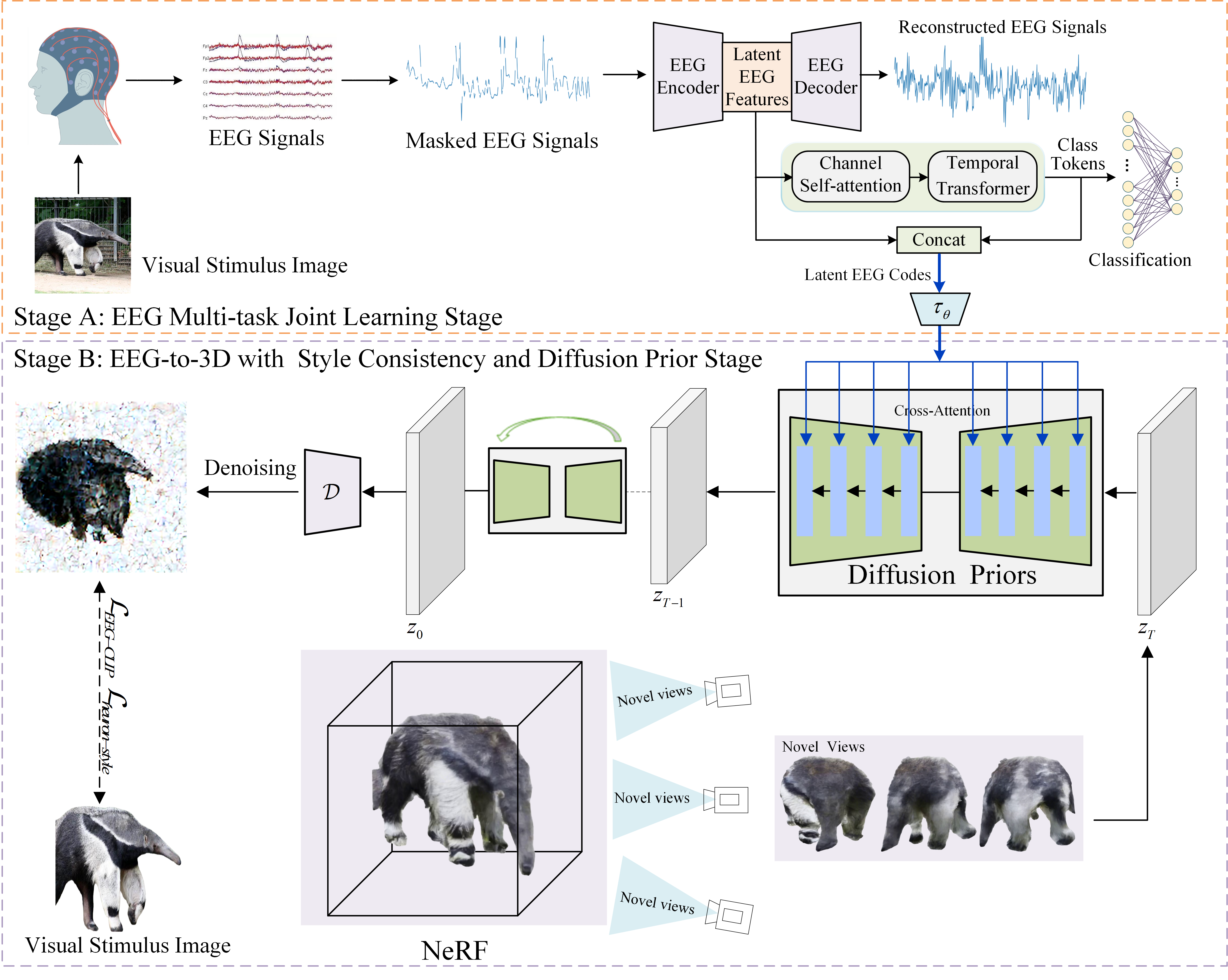} % image_file为要插入的图片文件名
    \caption{The overall network architecture of the proposed method.}
    \label{color_framework} % 添加图片引用标签
\end{figure*}
\section{The Neuroscientific Analysis of Our Method} \label{section3}
Our dataset is sourced from~\cite{spampinato2017deep}, where each image is displayed for 0.5 seconds, during which EEG is collected simultaneously. Based on references ~\cite{hegde2008time, fabre1998rapid, fabre2003visual, rayner1978eye}, it is known that the brain is capable of acquiring visual information within 0.5 seconds. Therefore, we hypothesize that EEG has already perceived specific 3D texture information within this 0.5-second window. Our work proposes this hypothesis and experimentally verifies its existence. To analyze how the brain captures visual perceptual information in such a short timeframe, we adopt a methodology incorporating 3D and color perception. This approach facilitates a more comprehensive explanation and understanding of the brain's visual perception process. Not only does it aid researchers in exploring perceptual mechanisms, but it also advances theoretical research pertaining to vision.

\subsection{3D Perception}
~\cite{korik2024real} conducted a series of experiments, including multiple participants who participated in two offline and three online sessions. These experiments successfully demonstrated the feasibility of distinguishing the neural correlates of imagined 3D objects in EEG. Specifically, they decoded the participants' imagined spheres, cones, pyramids, cylinders, and cubes.~\cite{yamane2008neural} discovered evidence of neural decoding for complex 3D object shapes. It utilized an evolutionary stimulus strategy and linear/nonlinear response models to characterize responses to 3D shapes. The configuration representation of 3D shapes can provide specific knowledge about object structures, supporting guidance for complex physical interactions.

\subsection{Color Perception}
% The study in reference ~\cite{liu2020hierarchical} shows that the human brain efficiently encodes visual information through intricately designed pathways, where color and direction are structurally interrelated in brain signals. This indicates that these visual features are not processed independently but are systematically encoded together. 
According to reference ~\cite{garg2019color}, using techniques such as intrinsic signal optical imaging, two-photon imaging, and EEG recording, the study detailed the hue map structures of different visual brain areas, revealing the neural mechanisms underlying the formation of color perception space. It was found that as the processing hierarchy of the visual cortex increases, the neurons encoding color in the brain gradually coordinate, ultimately forming a balanced mechanism that matches our subjective perception of hues.

These studies demonstrate that brain signals can perceive and process various visual information, such as color, shape, and texture.

\section{Methodology} \label{section4}
\subsection{Overview}
As shown in Figure \ref{color_framework}, to reconstruct style-consistent 3D objects from EEG signals, we propose an EEG-to-3D architecture with style consistency and semantic awareness. The proposed method mainly consists of two components: 1) a semantic-aware neural EEG encoder for 3D objects, and 2) a style-consistent neural decoder for 3D objects. First, we utilize EEG signals for neural encoding of 3D objects. In this stage, we obtain latent EEG features and class tokens through joint training tasks involving EEG signal reconstruction and classification, and these two components are merged and passed through a linear layer to generate latent EEG codes. Next, we use latent EEG codes obtained in the previous stage to provide conditional decoding features for the reverse diffusion process of the diffusion model via a cross-attention mechanism, generating novel 2D views while using content and style losses to ensure that the style of the novel 2D views remain as consistent as possible with GT. Finally, we use latent EEG codes to initialize LDM to generate novel 2D views from different perspectives to optimize NeRF, enabling the neural decoder to decode 3D objects.

\subsection{Stage A: Multi-Task Joint Learning}
During the initial phase, the process leverages EEG signals for neural encoding of 3D objects. Specifically, we first extract the regional semantic features of EEG by using a masked reconstruction model to capture the temporal features of EEG, thereby obtaining its regional information. Next, we perform a semantic classification task on EEG signals, enabling EEG encoder to learn semantic features. Through the EEG reconstruction task and semantic classification task, we capture the semantic features of regions, which serve as input for subsequent steps. Then, we explicitly introduce the semantic regional features of the original and reconstructed images by leveraging latent EEG codes containing latent information of semantic regions. We jointly fine-tune LDM to incorporate the performance of semantic regions, making the generated images more similar to GT in terms of regional and spatial relationships.

\subsubsection{Reconstruction and Classification Tasks of EEG}

\paragraph{\textbf{EEG Reconstruction Task}}Traditional models struggle to effectively extract meaningful features due to the complex spatial regional information characteristics of EEG. However, as demonstrated by the work and ideas of~\cite{chen2023seeing}, it is possible to capture valuable information from the context of EEG signals. Therefore, we use a masked model that randomly masks portions of EEG signals and then reconstructs them to achieve this purpose. Combining the findings of~\cite{poole2022dreamfusion} and~\cite{chen2023seeing}, we mask a certain proportion of EEG signals based on their temporal features and use~\cite{he2022masked} method to convert EEG into 1D data to embed it into the network. By considering the contextual temporal cues to predict the missing signals, we can learn the regional information.

\paragraph{\textbf{EEG-based Visual Classification Task}}Prior studies have reconstructed EEG signals using an EEG encoder ~\cite{chen2023seeing, BaiECCV24}. However, obtained latent EEG codes currently only encompass temporal and spatial characteristics, lacking crucial semantic features. Due to the lengthy temporal nature of EEG signals, establishing global relationships for EEG data classification poses a challenge. Therefore, we employ a Temporal Transformer to correlate time-series features with their own features, extracting global features of EEG over the time sequence. This allows for the extraction of temporal features from EEG signals, classification of EEG signals, and subsequent acquisition of class tokens. By integrating these features with the output from the EEG encoder, we aim to further enhance the accuracy of semantic classification.

\subsection{Stage B: Diffusion Prior and Style Consistency}
As Stage B, we utilize the reference view x generated by EEG and latent EEG codes to reconstruct NeRF, and constrain the novel view through a diffusion prior conditioned on the latent EEG codes. The proposed EEG-to-3D method is expected to simultaneously meet the following requirements: 1) It can generate 3D objects using EEG that closely resemble the rendering appearance of the reference view x; 2) Novel view renderings should exhibit semantic consistency with the reference view x, while also maintaining style consistency; 3) The generated 3D objects should possess well-defined geometric structures.

\paragraph{\textbf{Text-to-3D}}DreamFusion ~\cite{poole2022dreamfusion} demonstrated its capabilities in the field of text-to-3D synthesis by leveraging a pretrained text-to-image diffusion model ~\cite{saharia2022photorealistic} as a strong image prior. DreamFusion achieves text-to-3D generation through two key components: a pretrained text-to-image diffusion-based generative model and a neural scene representation of the scene model. The scene model is a parametric function $x=g(\theta )$, which generates an image $s$ at the specified camera pose. Here, $g$ is a volumetric renderer, and \(\theta\) is a coordinate-based multi-layer perceptron (MLP) representing a 3D volume. The diffusion model \(\phi\)  includes a learned denoising function ${{\epsilon }_{\theta }}({{x}_{t}};y,t)$ , which predicts the sampled noise \(\epsilon\) given the noisy image ${{x}_{t}}$, noise level $t$, and text embedding $y$. It provides the gradient direction to update \(\theta\) such that all rendered images are pushed toward high-probability density regions conditioned on the text embedding under the diffusion prior. Specifically, DreamFusion introduces Score Distillation Sampling (SDS) to compute the gradient.
\begin{equation}\label{SDS7}
    {{\nabla }_{\theta }}{{\mathcal{L}}_{SDS}}\left( \phi ,g\left( \theta  \right) \right)={{\mathbb{E}}_{t,\varepsilon }}\left[ \omega \left( t \right)\left( {{\epsilon }_{\theta }}\left( {{z}_{t}};y,t \right)-\epsilon  \right)\frac{\partial x}{\partial \theta } \right]
\end{equation}
Here, $\omega(t)$ is a weighting function. We regard the scene model $g$ and the diffusion model as modular components within the framework, with ${{\epsilon }_{\theta }}$ serving as the denoising function.

\paragraph{\textbf{EEG-to-3D with Diffusion Prior}} To ensure the semantic coherence of the EEG-to-3D objects, we adopt a diffusion prior to impose additional constraints on the novel view renderings. Previous works on text-to-3D generation ~\cite{poole2022dreamfusion,tang2023make} have applied ${{\mathcal{L}}_{SDS}}$ to leverage text-conditioned diffusion models as 2D diffusion priors. In the method proposed in this paper, we employ a diffusion prior conditioned on latent EEG codes to optimize the views generated by NeRF, gradually refining them from blurry to sharp. In this case, to apply ${{\mathcal{L}}_{SDS}}$, we use latent EEG codes obtained in the first stage as the latent prompt $y$, allowing us to perform ${{\mathcal{L}}_{SDS}}$ within the latent space of the diffusion model: 
\begin{equation}\label{SDS8}
    {{\nabla }_{\theta }}{{\mathcal{L}}_{SDS}}\left( \phi ,g\left( \theta  \right) \right)={{\mathbb{E}}_{t,\varepsilon }}\left[ \omega \left( t \right)\left( {{\epsilon }_{\theta }}\left( {{z}_{t}};y,t \right)-\epsilon  \right)\frac{\partial z}{\partial x}\frac{\partial x}{\partial \theta } \right]
\end{equation}
here, $y$ represents latent EEG codes, ${{z}_{t}}$ denotes the noisy latent representation, 
$\frac{\partial z}{\partial x}$ refers to the gradient of the LDM encoder, and $\frac{\partial x}{\partial \theta }$ represents the gradient of the rendered image.

\paragraph{\textbf{EEG-Image Information Aligning with NeRF}}According to ${{\mathcal{L}}_{SDS}}$ in DreamFusion, this function is used to measure the similarity between novel views and text prompts. However, in this work, 3D objects are generated based on EEG signals, thus requiring further alignment between the EEG signals and novel views from different perspectives. To achieve this, we leverage a pretrained CLIP image encoder to align the EEG signals with the novel views. The corresponding loss function is as follows:
\begin{equation}\label{EEG-Image}
    {{\mathcal{L}}_{EEG\\-CLIP}}=-{{E}_{CLIP}}(T/I)\cdot \rho ({{\varphi }_{latent}}(y))\cdot {{E}_{CLIP}}(g(\theta ))
\end{equation}
Here, ${{E}_{CLIP}}(T/I)$ includes the CLIP image encoder. By adding a similarity loss to the novel view image $g(\theta )$ obtained under the diffusion prior of EEG, our goal is to align the novel view image with the reference image.

\begin{table*}[hbt!]
    \small
    \centering
    % table caption is above the table
    % For LaTeX tables use
    \begin{tabular}{cccccc}
        \hline
        Visual Stimulus Images   &w/o ${{\mathcal{L}}_{neuron-style}}$  &with ${{\mathcal{L}}_{style}}$  & 3D Views (${20}^{o}$)   &${45}^{o}$  &${70}^{o}$ \\
        \hline
        \makecell[c]{\\
        \begin{minipage}[b]{0.22\columnwidth}
            \centering
            {\includegraphics[width=0.8\textwidth]{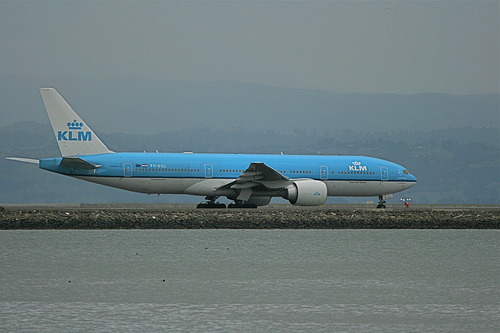}}
        \end{minipage}
        }
        &\makecell[c]{\\
        \begin{minipage}[b]{0.22\columnwidth}
            \centering
            {\includegraphics[width=0.8\textwidth]{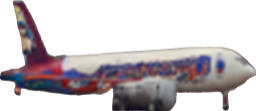}}
        \end{minipage}
        }  
        &\makecell[c]{\\
        \begin{minipage}[b]{0.22\columnwidth}
            \centering
            {\includegraphics[width=0.7\textwidth]{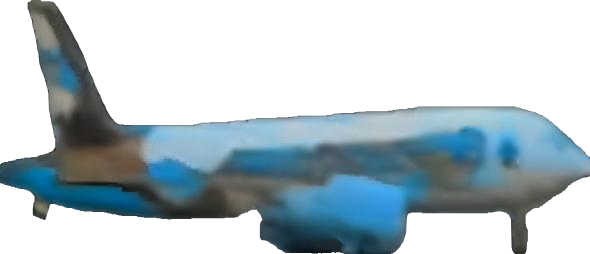}}
        \end{minipage}
        } 
        & \makecell[c]{\\
        \begin{minipage}[b]{0.22\columnwidth}
            \centering
            {\includegraphics[width=0.7\textwidth]{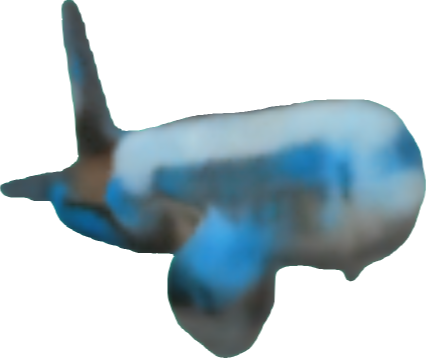}}
        \end{minipage}
        }
        & \makecell[c]{\\
        \begin{minipage}[b]{0.22\columnwidth}
            \centering
            {\includegraphics[width=0.7\textwidth]{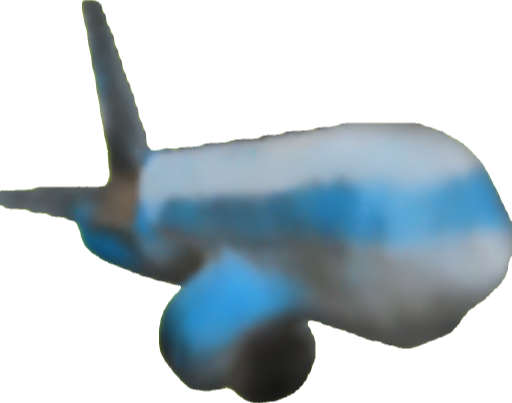}}
        \end{minipage}
        }
        & \makecell[c]{\\
        \begin{minipage}[b]{0.22\columnwidth}
            \centering
            {\includegraphics[width=0.7\textwidth]{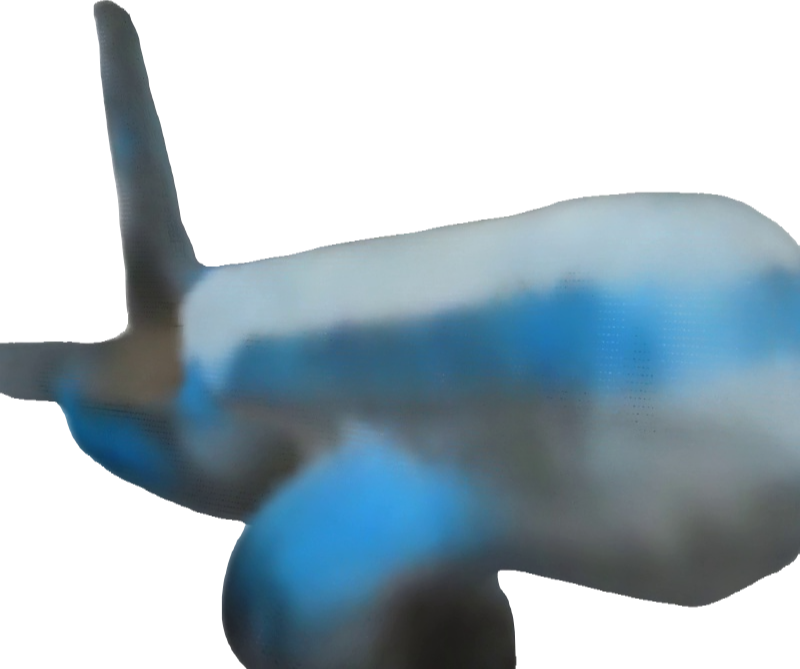}}
        \end{minipage}
        }
        \\

        % \makecell[c]{\\
        % \begin{minipage}[b]{0.22\columnwidth}
        %     \centering
        %     {\includegraphics[width=0.8\textwidth]{ma_gt.JPEG}}
        % \end{minipage}
        % }
        % & \makecell[c]{\\
        % \begin{minipage}[b]{0.22\columnwidth}
        %     \centering
        %     {\includegraphics[width=0.8\textwidth]{ma_no_color.png}}
        % \end{minipage}
        % } 
        % & \makecell[c]{\\
        % \begin{minipage}[b]{0.22\columnwidth}
        %     \centering
        %     {\includegraphics[width=0.7\textwidth]{ma_pre_color.png}}
        % \end{minipage}
        % }
        % & \makecell[c]{\\
        % \begin{minipage}[b]{0.22\columnwidth}
        %     \centering
        %     {\includegraphics[width=0.7\textwidth]{output_ma_20.png}}
        % \end{minipage}
        % }
        % & \makecell[c]{\\
        % \begin{minipage}[b]{0.22\columnwidth}
        %     \centering
        %     {\includegraphics[width=0.7\textwidth]{output_ma_45.png}}
        % \end{minipage}
        % }
        % & \makecell[c]{\\
        % \begin{minipage}[b]{0.22\columnwidth}
        %     \centering
        %     {\includegraphics[width=0.7\textwidth]{output_ma_70.png}}
        % \end{minipage}
        % }
        % \\

       \makecell[c]{\\
        \begin{minipage}[b]{0.22\columnwidth}
            \centering
            {\includegraphics[width=0.8\textwidth]{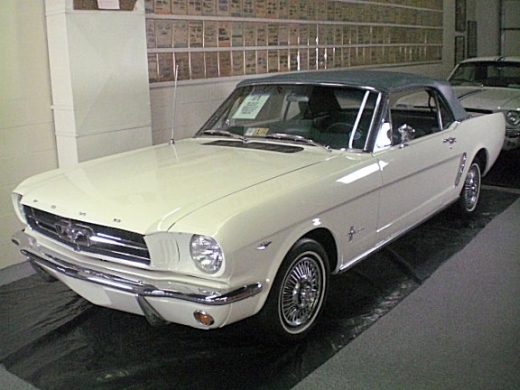}}
        \end{minipage}
        }
        & \makecell[c]{\\
        \begin{minipage}[b]{0.22\columnwidth}
            \centering
            {\includegraphics[width=0.8\textwidth]{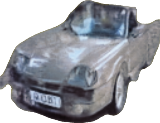}}
        \end{minipage}
        }
        & \makecell[c]{\\
        \begin{minipage}[b]{0.22\columnwidth}
            \centering
            {\includegraphics[width=0.7\textwidth]{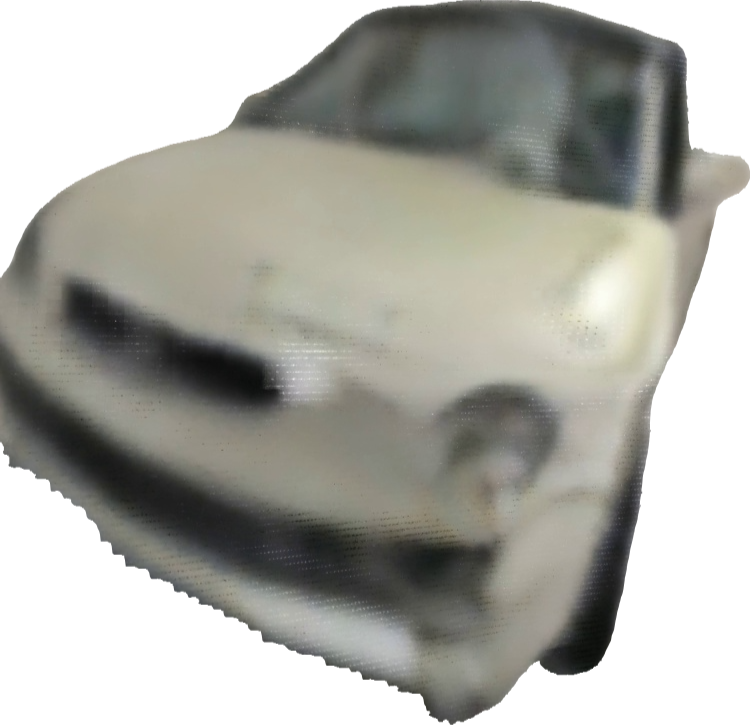}}
        \end{minipage}
        }
        & \makecell[c]{\\
        \begin{minipage}[b]{0.22\columnwidth}
            \centering
            {\includegraphics[width=0.7\textwidth]{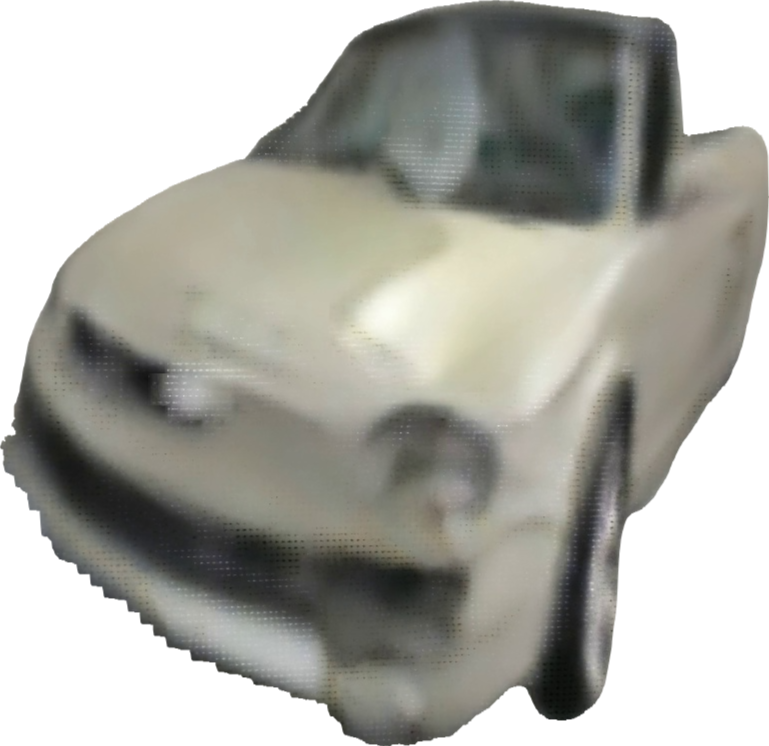}}
        \end{minipage}
        }
        & \makecell[c]{\\
        \begin{minipage}[b]{0.22\columnwidth}
            \centering
            {\includegraphics[width=0.7\textwidth]{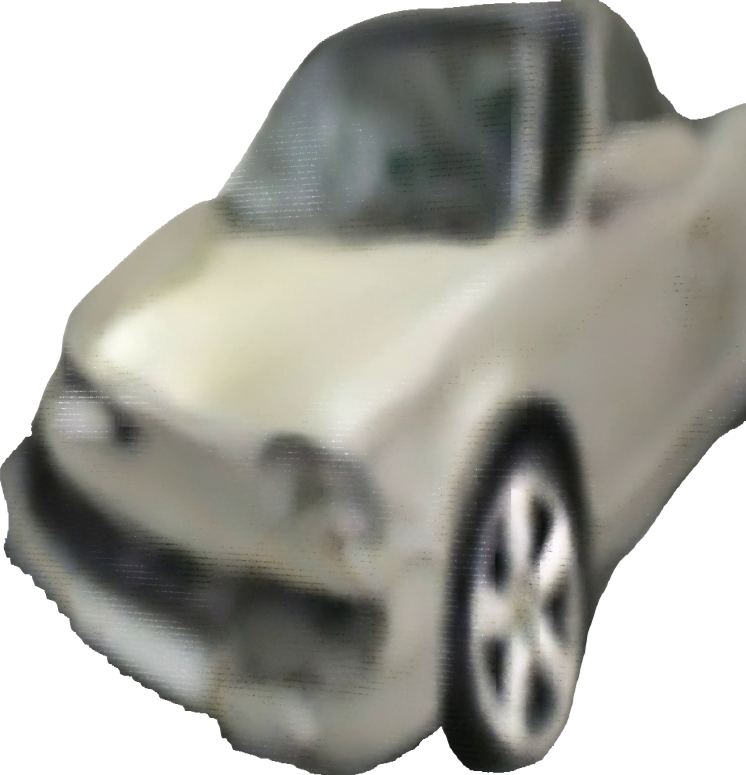}}
        \end{minipage}
        }
        & \makecell[c]{\\
        \begin{minipage}[b]{0.22\columnwidth}
            \centering
            {\includegraphics[width=0.7\textwidth]{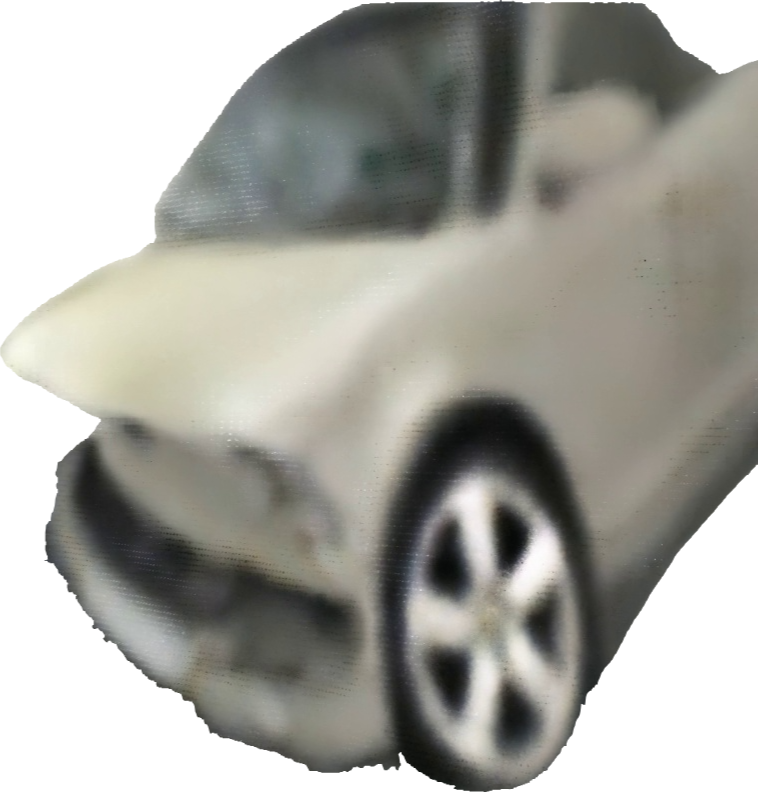}}
        \end{minipage}
        }
        \\ 
        \hline
    \end{tabular}
    \caption{Typical 3D objects generated by EEG-driven models based on the EEG-ImageNet dataset ~\cite{spampinato2017deep}.}
    \label{table_cvpr2027}
\end{table*}

\begin{table*}[hbt!]
    \small
    \centering
    % table caption is above the table
    % For LaTeX tables use
    \begin{tabular}{ccccc}
        \hline
        Visual Stimulus Images     &with ${{\mathcal{L}}_{neuron-style}}$  & 3D Views (${20}^{o}$)   &${45}^{o}$  &${70}^{o}$ \\
        \hline
        \makecell[c]{\\
        \begin{minipage}[b]{0.2\columnwidth}
            \centering
            {\includegraphics[width=0.8\textwidth]{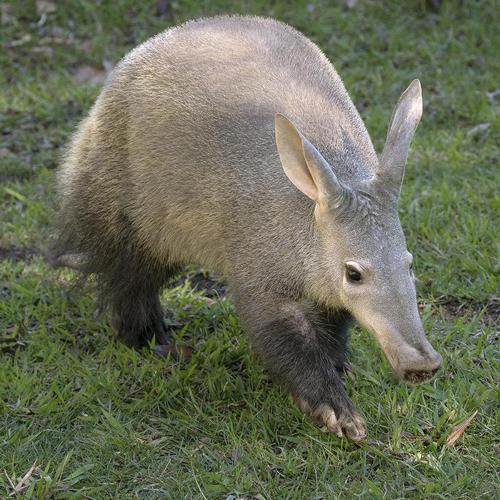}}
        \end{minipage}
        }
        &\makecell[c]{\\
        \begin{minipage}[b]{0.2\columnwidth}
            \centering
            {\includegraphics[width=1\textwidth]{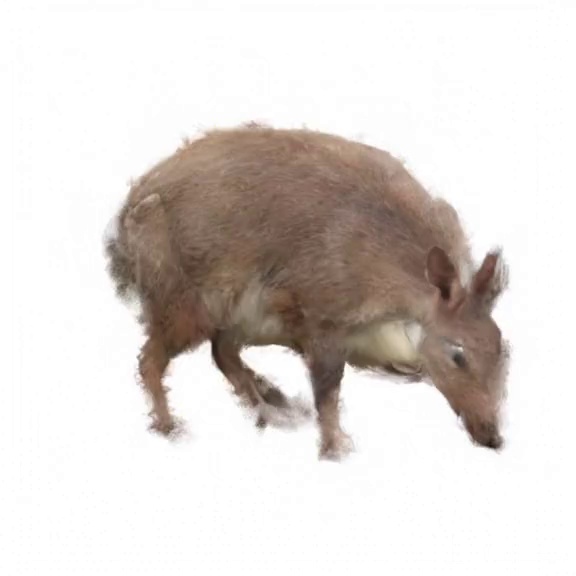}}
        \end{minipage}
        }  
        &\makecell[c]{\\
        \begin{minipage}[b]{0.2\columnwidth}
            \centering
            {\includegraphics[width=1\textwidth]{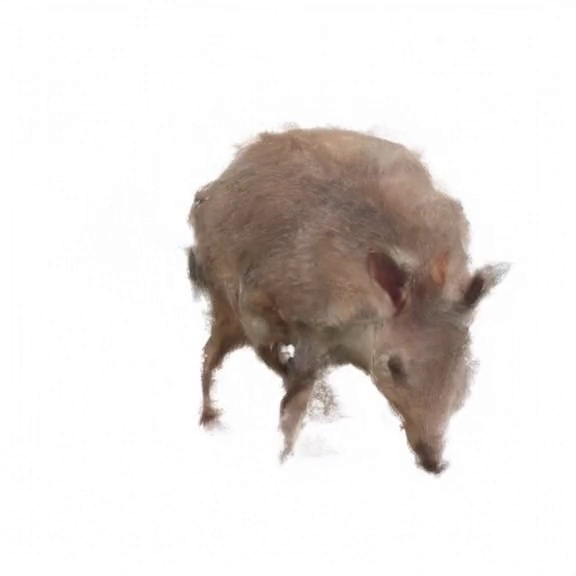}}
        \end{minipage}
        } 
        & \makecell[c]{\\
        \begin{minipage}[b]{0.2\columnwidth}
            \centering
            {\includegraphics[width=1\textwidth]{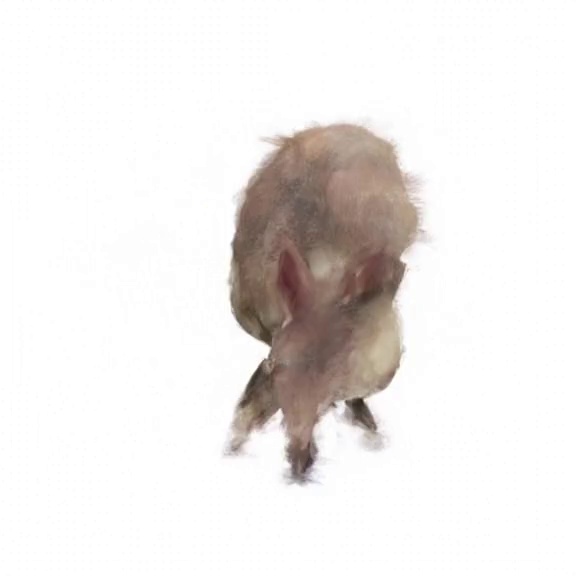}}
        \end{minipage}
        }
        & \makecell[c]{\\
        \begin{minipage}[b]{0.2\columnwidth}
            \centering
            {\includegraphics[width=1\textwidth]{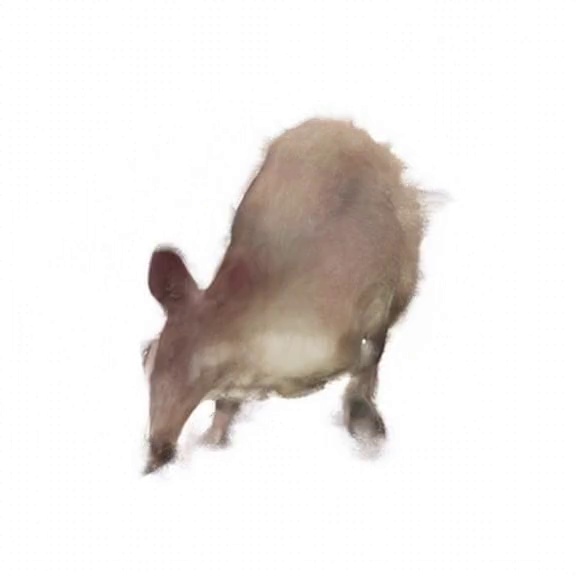}}
        \end{minipage}
        }
        \\
       \makecell[c]{\\
        \begin{minipage}[b]{0.2\columnwidth}
            \centering
            {\includegraphics[width=0.8\textwidth]{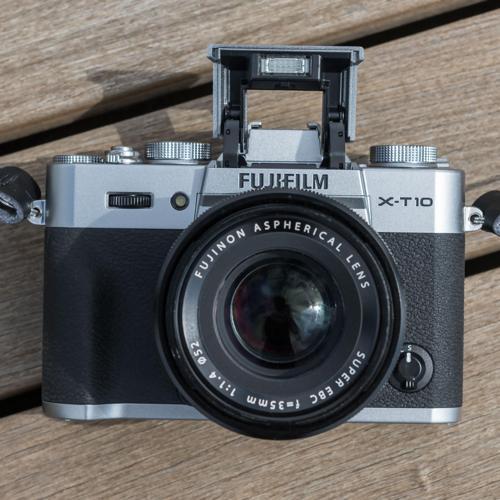}}
        \end{minipage}
        }
        &\makecell[c]{\\
        \begin{minipage}[b]{0.2\columnwidth}
            \centering
            {\includegraphics[width=1\textwidth]{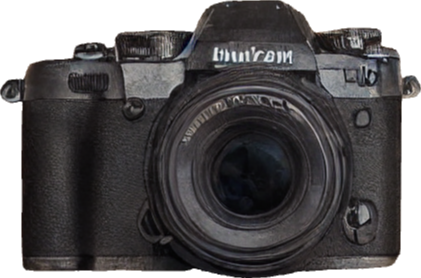}}
        \end{minipage}
        }  
        &\makecell[c]{\\
        \begin{minipage}[b]{0.2\columnwidth}
            \centering
            {\includegraphics[width=1\textwidth]{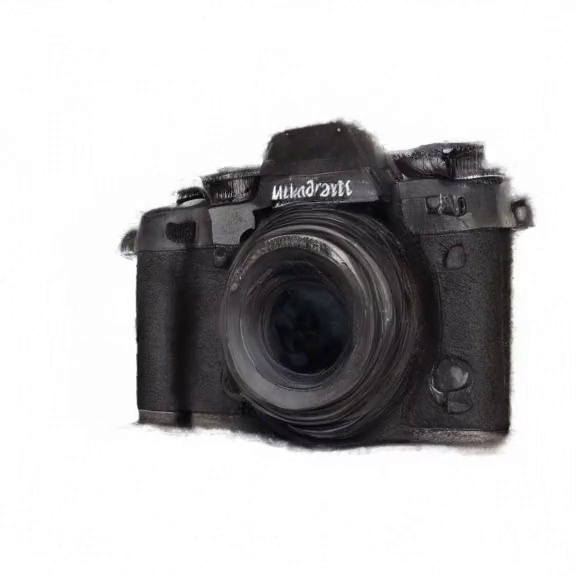}}
        \end{minipage}
        } 
        & \makecell[c]{\\
        \begin{minipage}[b]{0.2\columnwidth}
            \centering
            {\includegraphics[width=1\textwidth]{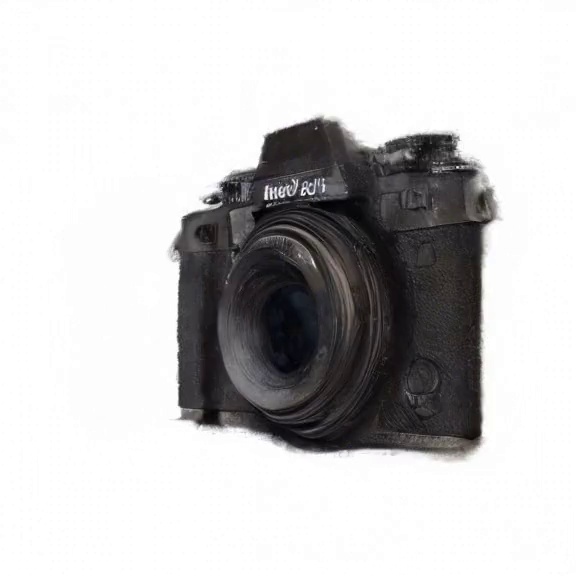}}
        \end{minipage}
        }
        & \makecell[c]{\\
        \begin{minipage}[b]{0.2\columnwidth}
            \centering
            {\includegraphics[width=1\textwidth]{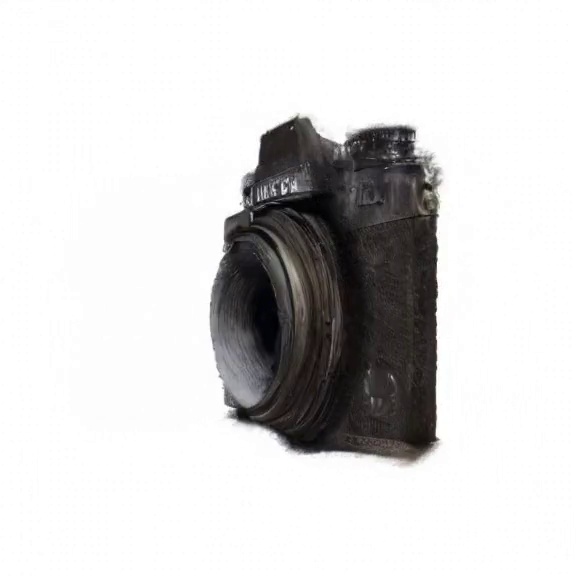}}
        \end{minipage}
        }
        \\
        
       \hline
        \makecell[c]{\\
        \begin{minipage}[b]{0.2\columnwidth}
            \centering
            {\includegraphics[width=1\textwidth]{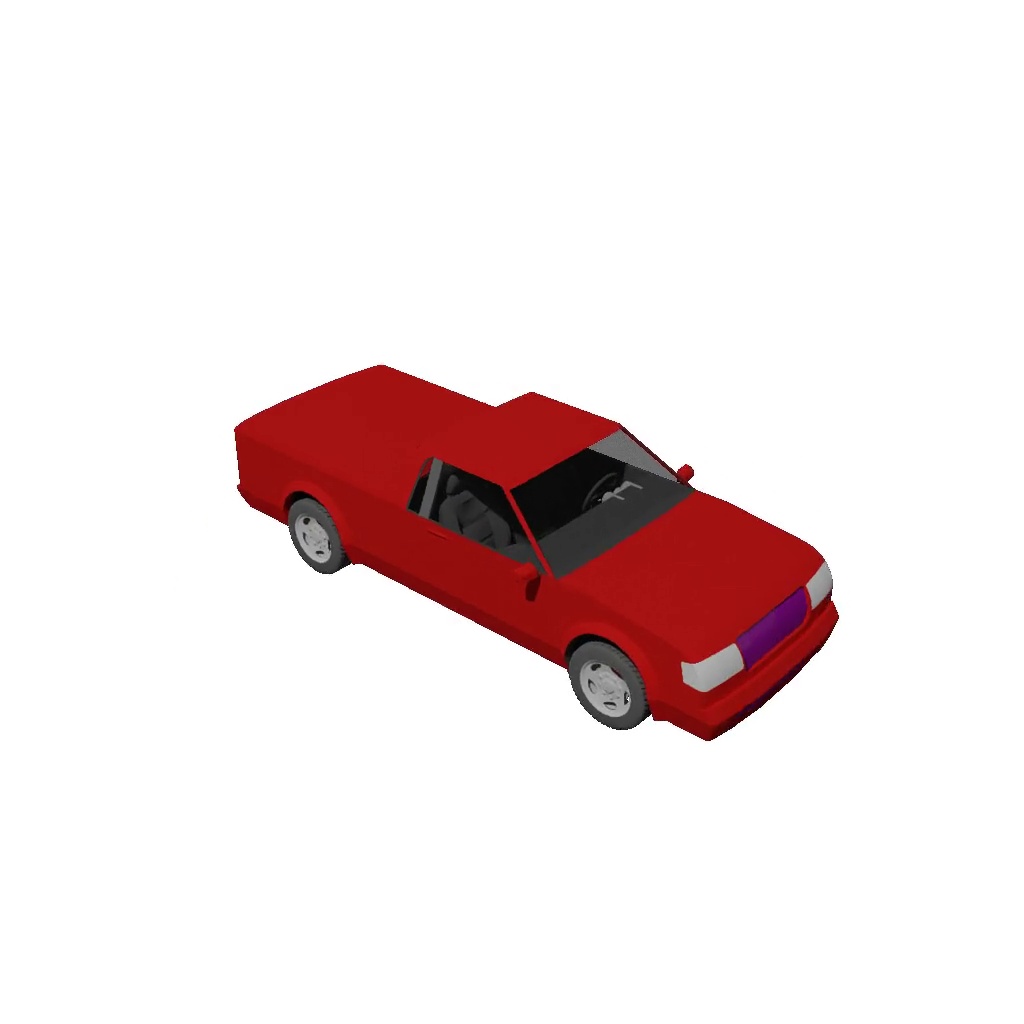}}
        \end{minipage}
        }
        &\makecell[c]{\\
        \begin{minipage}[b]{0.2\columnwidth}
            \centering
            {\includegraphics[width=0.5\textwidth]{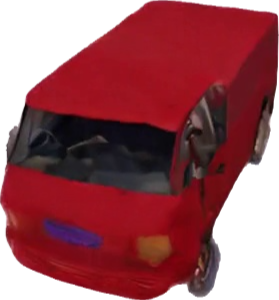}}
        \end{minipage}
        }  
        &\makecell[c]{\\
        \begin{minipage}[b]{0.2\columnwidth}
            \centering
            {\includegraphics[width=0.7\textwidth]{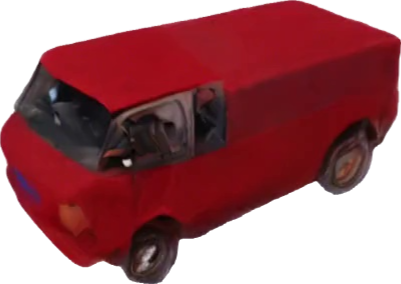}}
        \end{minipage}
        } 
        & \makecell[c]{\\
        \begin{minipage}[b]{0.2\columnwidth}
            \centering
            {\includegraphics[width=0.7\textwidth]{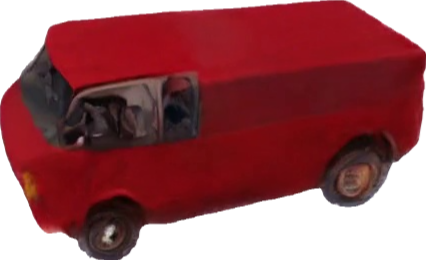}}
        \end{minipage}
        }
        & \makecell[c]{\\
        \begin{minipage}[b]{0.2\columnwidth}
            \centering
            {\includegraphics[width=0.7\textwidth]{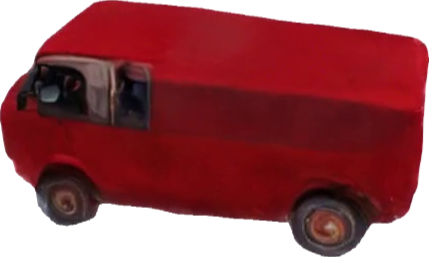}}
        \end{minipage}
        }        
        \\        
        \makecell[c]{\\
        \begin{minipage}[b]{0.2\columnwidth}
            \centering
            {\includegraphics[width=0.7\textwidth]{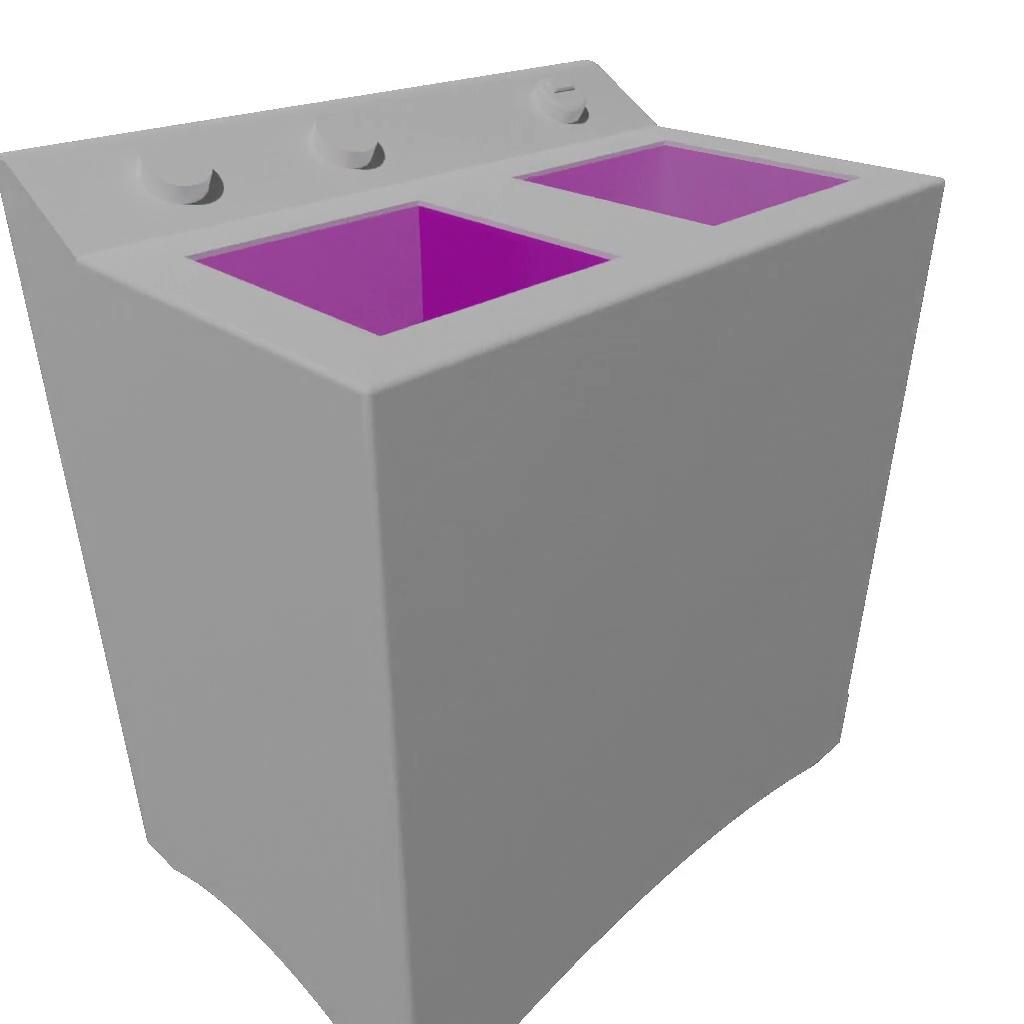}}
        \end{minipage}
        }
        &\makecell[c]{\\
        \begin{minipage}[b]{0.2\columnwidth}
            \centering
            {\includegraphics[width=0.7\textwidth]{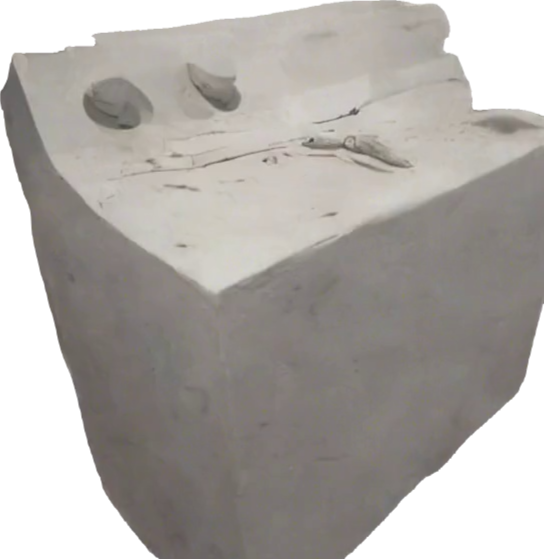}}
        \end{minipage}
        }  
        &\makecell[c]{\\
        \begin{minipage}[b]{0.2\columnwidth}
            \centering
            {\includegraphics[width=0.7\textwidth]{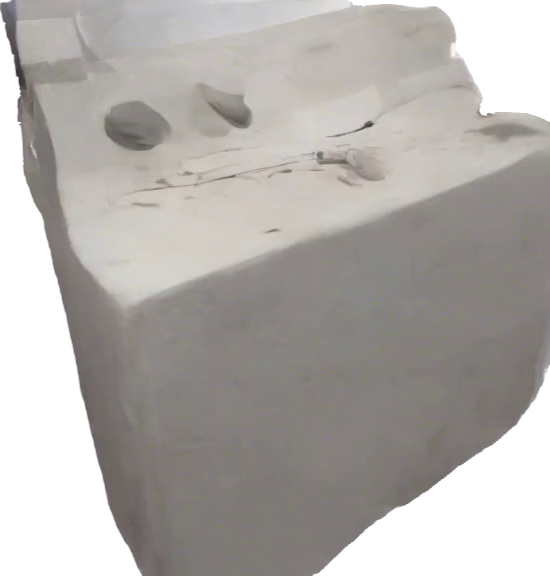}}
        \end{minipage}
        } 
        & \makecell[c]{\\
        \begin{minipage}[b]{0.2\columnwidth}
            \centering
            {\includegraphics[width=0.7\textwidth]{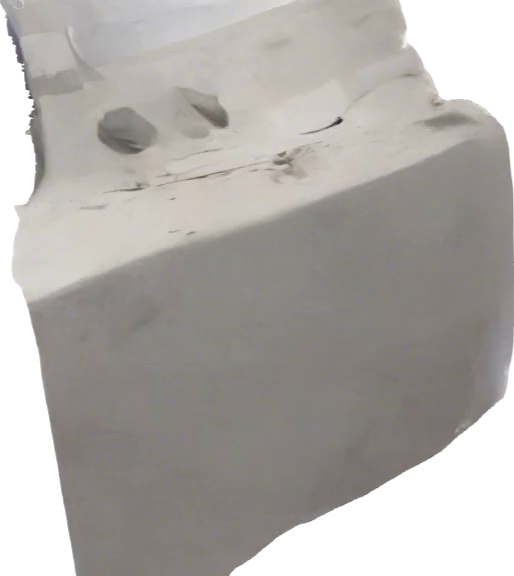}}
        \end{minipage}
        }
        & \makecell[c]{\\
        \begin{minipage}[b]{0.2\columnwidth}
            \centering
            {\includegraphics[width=0.7\textwidth]{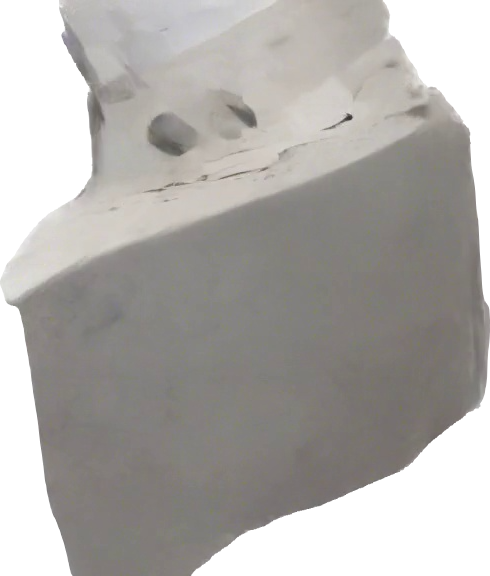}}
        \end{minipage}
        }
        \\   
        % Latent EEG Codes &\makecell[c]{\\
        % \begin{minipage}[b]{0.2\columnwidth}
        %     \centering
        %     {\includegraphics[width=1\textwidth]{yizi_gt.JPEG}}
        % \end{minipage}
        % }
        % &\makecell[c]{\\
        % \begin{minipage}[b]{0.2\columnwidth}
        %     \centering
        %     {\includegraphics[width=1\textwidth]{yizi_color.jpg}}
        % \end{minipage}
        % }  
        % &\makecell[c]{\\
        % \begin{minipage}[b]{0.2\columnwidth}
        %     \centering
        %     {\includegraphics[width=1\textwidth]{yizi_color_20.jpg}}
        % \end{minipage}
        % } 
        % & \makecell[c]{\\
        % \begin{minipage}[b]{0.2\columnwidth}
        %     \centering
        %     {\includegraphics[width=1\textwidth]{yizi_color_45.jpg}}
        % \end{minipage}
        % }
        % & \makecell[c]{\\
        % \begin{minipage}[b]{0.2\columnwidth}
        %     \centering
        %     {\includegraphics[width=1\textwidth]{yizi_color_70.jpg}}
        % \end{minipage}
        % }
        % \\
        \hline
    \end{tabular}
    \caption{The method proposed in this study reconstructs 3D objects with typical style consistency using the Things-EEG2 ~\cite{gifford2022large} and fMRI-Image ~\cite{gao2023mind} datasets, respectively.}
    \label{eeg2_3d}
\end{table*}

\begin{figure}
    \centering
    \includegraphics[width=0.9\linewidth]{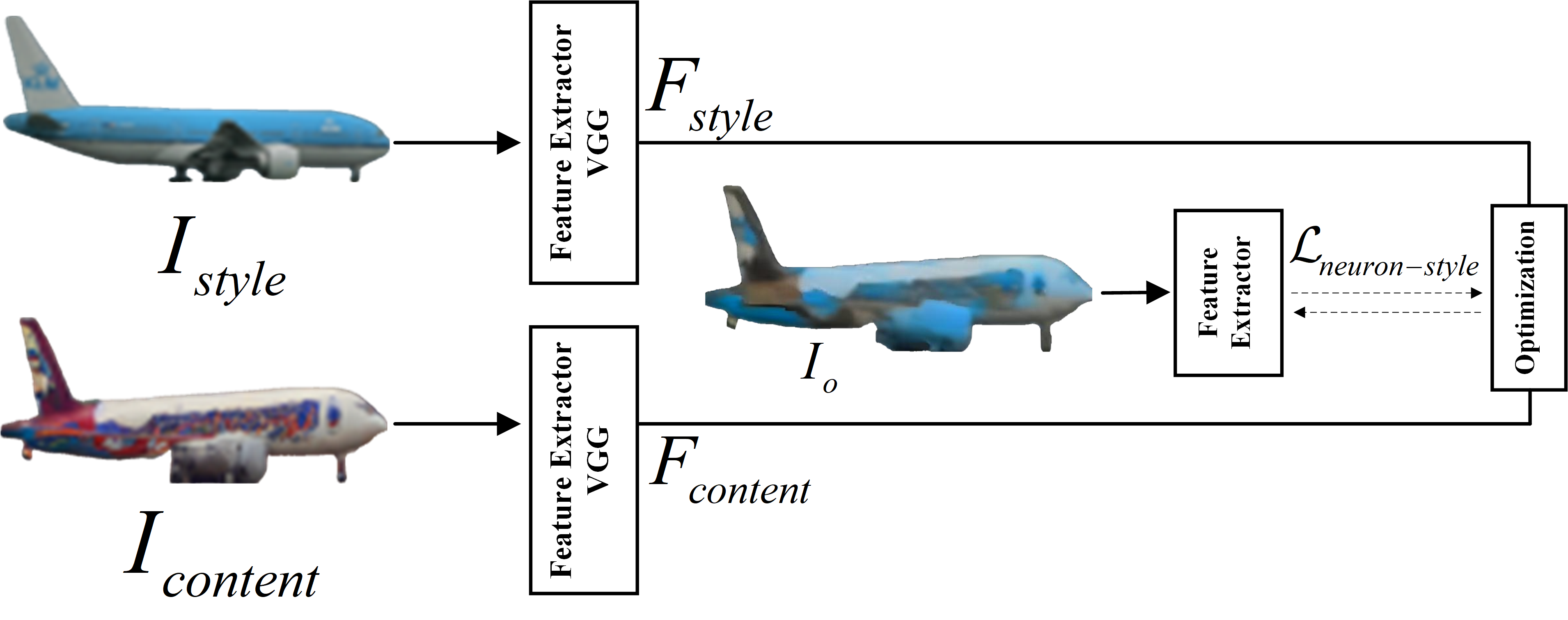}
    \caption{Neural style transfer. In Stage B of the proposed method, a color transfer loss is incorporated.}
    \label{style_method}
\end{figure}

\paragraph{\textbf{Neural Style Transfer}}As shown in Figure \ref{style_method}, the goal of neural style transfer is to transfer the style ${{I}_{s}}$ of GT onto the content image ${{I}_{c}}$, generating a stylized image ${{I}_{o}}$, while ensuring that the reconstructed image preserves the color realism of the GT and the content remains unchanged. According to ~\cite{Chiu_2022_CVPR, Li_2018_ECCV}, adding a new loss term to the optimization objective enhances the realism of stylized images produced by the neural style transfer algorithm, particularly in preserving the local structures of the content image. Building upon these references, we incorporate both style loss and content loss in the style transfer process. We utilize a pretrained VGG network to extract features from the target image, which are used to compute the content and style losses. By combining these two losses, the network generates new images that retain the content while adopting a specific style. To achieve the aforementioned style transfer, we use the pretrained VGG feature extractor $\phi (\cdot )$ to extract content and style features, denoted as ${{F}_{c}}=\phi ({{I}_{c}})$ and ${{F}_{s}}=\phi ({{I}_{s}})$, respectively. The output image ${{I}_{o}}$ is optimized using the following objective function, which includes content loss ${{\mathcal{L}}_{content}}$ and style loss ${{\mathcal{L}}_{style}}$:
\begin{equation}\label{style}
{{\mathcal{L}}_{neuron-style}}=argmin\{{{\mathcal{L}}_{content}}(\varsigma (I),{{F}_{c}})+\lambda {{\mathcal{L}}_{style}}(\varsigma (I),{{F}_{s}})\}
\end{equation}
here, $\varsigma (\cdot )$ represents the feature extractor, and $\lambda$ is the balancing factor between the content loss and style loss.

\paragraph{\textbf{Loss Function with Diffusion Prior and Style Consistency}}The overall loss in the second stage can be represented by ${{\mathcal{L}}_{SDS}}$, ${{\mathcal{L}}_{EEG-CLIP}}$, and ${{\mathcal{L}}_{neuron-style}}$. In this work, the diffusion prior generated under the condition of latent EEG codes is used to guide NeRF in generating 3D objects with consistent style. ${{\mathcal{L}}_{SDS}}$ is used to optimize NeRF to achieve better geometric shapes and details, ${{\mathcal{L}}_{EEG-Image}}$ is employed to align the EEG, text, and image, and ${{\mathcal{L}}_{style}}$ is used to adjust the reconstructed style of the 3D objects to ensure they remain as consistent as possible with GT color, thereby enhancing the realism and visual accuracy of the 3D reconstruction.

\paragraph{\textbf{Loss Function between EEG and Latent Diffusion Models}} Under the combined influence of reconstruction and classification based on EEG signals, an EEG encoder with temporal, spatial, and semantic features is obtained. Using this EEG encoder along with semantic embedding, we map EEG signals to a latent space constrained by semantic regions. We integrate cross-attention mechanisms and latent conditional features $t$, and utilize the denoising U-Net to replace the original temporal pixel space features in LDM with conditional latent features ${{z}_{t}}$ , thereby optimizing the loss function. This forms the conditional loss function for LDM:
\begin{equation}\label{ldm4}
    {{\mathcal{L}}_{ldm}} = {E_{z,\varepsilon \sim~N(0,1),t}}\left[ {\left\| {\varepsilon  - {\varepsilon _\theta }({z_t},t,{\tau _\theta }(y))} \right\|_2^2} \right]\
\end{equation}

To endow LDM with regional semantic performance, we utilize the pre-trained Segment Anything (SAM) ~\cite{kirillov2023segment} to obtain the regional semantic maps of visual stimulus images and their corresponding reconstructed images. The regional semantic loss is calculated by the cross-entropy loss function as follows,
\begin{equation}\label{ldm_region}
  {{\mathcal{L}}_{region}} = -\frac{1}{N}
  \sum\limits_{i=1}^N
  {\sum\limits_{k=1}^M
  {p_{i,k}}\cdot\log\left(\hat{p}_{i,k}\right)}
\end{equation}
where $N$ is the number of pixels, $M$ is the number of categories, $p=\mathrm{SAM}\left(S\right)$ and $\hat{p}=\mathrm{SAM}\left(S'\right)$ denote the regional semantic maps of the visual stimulus image $S$ and its corresponding reconstructed image ${S}'$, respectively. Combining \eqref{ldm4} and \eqref{ldm_region}, we finally obtain:
\begin{equation}\label{con:ldm_total}
    {{\mathcal{L}}_{ldm-region}} = {\lambda _{ldm}}{{\mathcal{L}}_{ldm}} + {\lambda _{region}}{{\mathcal{L}}_{region}},
\end{equation}
where ${{\lambda }_{ldm}}$ and ${{\lambda }_{region}}$ are balancing factors used to balance the influence between ${{\mathcal{L}}_{ldm}}$ and ${{\mathcal{L}}_{region}}$, with default values of 1. Subsequently, through backpropagation, we continuously fine-tune and update the weights of the LDM to endow it with the capability to capture semantic regions.

\section{Experiments}
\subsection{Dataset and Implementation Details}
% To validate that the method proposed in this paper can reconstruct style-consistent 3D objects using brain signals, we conducted extensive experiments on the EEG-ImageNet ~\cite{spampinato2017deep}, Things-EEG2 ~\cite{gifford2022large}, and fMRI-Image ~\cite{gao2023mind} datasets. 

\paragraph{\textbf{EEG-ImageNet Dataset}}The EEG-ImageNet dataset comes from the PeRCeiVe Lab ~\cite{spampinato2017deep}, which contains visual stimulus EEG signals recorded from 6 subjects. The visual stimulus images encompass a total of 40 categories, with each category comprising 50 images sourced from ImageNet. Each image within the same category is displayed consecutively for 0.5 seconds, and there is a 10-second interval between each category. The collected dataset comprises a total of 11,964 segments (corresponding to visual stimuli for EEG). Each EEG segment includes 128 channels, and the EEG signals are in three frequency ranges: 14-70Hz, 5-95Hz, and 55-95Hz. 

\paragraph{\textbf{Things-EEG2 Dataset}}To validate the effectiveness of the proposed method, we introduce an additional Things-EEG2 dataset ~\cite{gifford2022large}. This dataset includes ten subjects, with a total of 1,654 training sample categories. Each image is displayed for 100 ms with a 750 ms blank interval. All participants completed four equivalent experiments, resulting in 16,540 training images, each condition repeated four times, and 200 test images, each condition repeated 80 times.

\paragraph{\textbf{fMRI-Image Dataset}}The fMRI dataset comes from Mind-3D ~\cite{gao2023mind}, and in the supplementary materials, the methods proposed in this paper are compared with Mind-3D. The methods in this paper reconstruct 3D objects with consistent style.

% \paragraph{\textbf{Implementation Details of EEG Encoder}}The hardware configuration for the proposed method utilizes an NVIDIA A100 GPU. At the current stage, the embedding size of the encoder's pretrained model is set to 1024, with an encoder depth of 24, and the masking rate is typically 0.75. The length of the EEG signals is standardized to 512, and the optimizer chosen for gradient descent is Adam. These basic parameters are consistent with the configuration of the pretrained LDM. After instance segmentation, the original and reconstructed images are processed into semantic images with a resolution of 512x512. During the LDM training process, images generated after 300 epochs are presented at a resolution of 512x512.

\paragraph{\textbf{Implementation Details of EEG-to-3D}}In this work, the latent EEG codes obtained in Stage A are used to guide LDM in generating diffusion priors, with Stable Diffusion version 1.5 employed. To accelerate NeRF training and rendering in Stage B, the proposed method adopts Instant-NGP ~\cite{muller2022instant}. We use ${{\mathcal{L}}_{SDS}}$, ${{\mathcal{L}}_{EEG-Image}}$, and ${{\mathcal{L}}_{neuron-style}}$ to compute the loss for generating 3D objects, and the Adam optimizer is used to update the model parameters, with the learning rate set to 0.001. Our hardware configuration utilizes an NVIDIA A100 GPU

\begin{table*}
  \small %footnotesize
  %\color{blue}
  \centering
  \begin{tabular}{lccccc}
    \hline
    Models & Acc. & FID ~\cite{FID} & IS ~\cite{Szegedy} & Size (GB) & Inference time\\
    \hline
    EEGStyleGAN-ADA ~\cite{singh2024learning} & 0.38 & 10.82 & \diagbox{}{} & 0.458 & \diagbox{}{} \\
    DreamDiffusion~\cite{BaiECCV24} & 0.46 & \diagbox{}{} & \diagbox{}{} &5.4 &50 s/pic\\
    EEG-Decoding~\cite{ferrante2024decoding}  & 0.47 & \diagbox{}{} & \diagbox{}{} & \diagbox{}{} & \diagbox{}{} \\
    EEGVis-CMR~\cite{ye2024self}    & 0.51 & \diagbox{}{} & \diagbox{}{} & \diagbox{}{} & \diagbox{}{} \\
    Brain2Image~\cite{KavasidisMM17} & \diagbox{}{} & \diagbox{}{} & 5.01   & \diagbox{}{}  & \diagbox{}{} \\
    NeuroVision~\cite{Khare} & \diagbox{}{} & \diagbox{}{} & 5.23  & \diagbox{}{}  & \diagbox{}{} \\
    \bf{Ours} & \bf{0.67} & \bf{2.17} & \bf{25.41} & \bf{7.8} &\bf{50 s/pic} \\
    \hline
    \end{tabular}
      \caption{Quantitative metrics for 2D image reconstruction.}
    \label{Quantitative_2d}
\end{table*}

\begin{table*}[hbt!]
  % \small %footnotesize
  \centering
  \begin{tabular}{cccccc}
    \hline
     & Average Value & ${{0}^{o}}$ &${{20}^{o}}$ &${{45}^{o}}$ &${{70}^{o}}$\\
    \hline
    LPIPS ~\cite{zhang2018unreasonable}      & 0.3384 & 0.2403 & 0.3617 & 0.4539 & 0.4952 \\
    Contextual ~\cite{mechrez2018contextual} & 2.7943 & 2.3406 & 2.9274 & 3.1158 & 3.2932\\
    Acc ~\cite{radford2021learning}          & 0.9468 & 0.8259 & 0.8664 & 0.8093 & 0.8917 \\
    \hline
    \end{tabular}
    \caption{Quantitative LPIPS and Contextual metrics for different viewpoints of 3D objects. Acc refers to the scores obtained using CLIP. This study measures the results by evaluating the semantic similarity between the generated views and the reference views.}
    \label{table3_3d}
\end{table*}

\begin{table*}[hbt!] 
  \small %footnotesize
  \centering
  \begin{tabular}{ccccc}
    \hline
     & Full & w/o ${{\mathcal{L}}_{neuron-style}}$ &w/o ${{\mathcal{L}}_{EEG-CLIP}}$ &w/o ${{L}_{region}}$\\
    \hline
    LPIPS       & 0.3384 & 0.5981 & 0.3061 & 0.5827  \\
    Contextual  & 2.7943 & 3.8915 & 3.2685 & 3.6412  \\
    Acc         & 0.9468 & 0.8134 & 0.5837 & 0.7258  \\
    \hline
    \end{tabular}
    \caption{A quantitative comparison of different design choices from the perspectives of LPIPS and Contextual metrics.}
    \label{Ablation Study 3d}
\end{table*}
\subsection{Results}
% \subsubsection{Quality with 2D}
% As illustrated in Table~\ref{table_cvpr2027}, in Stage A, this study leverages EEG data combined with fine-tuning of LDM to reconstruct the generated 2D images. By accurately decoding the latent space that encapsulates semantic region information through EEG, the fine-tuned LDM successfully achieves a high-fidelity reconstruction of the original images.

\subsubsection{Quality with 3D and Style Consistency}
As shown in Table~\ref{table_cvpr2027} and ~\ref{eeg2_3d}, this paper reconstructs 3D objects from EEG signals under two conditions: with and without style transfer loss. For the reconstruction of 3D objects with style consistency, the proposed method integrates latent EEG codes, semantically accurate reconstructed images, and style transfer to generate 3D objects with consistent style. These reconstructed 3D objects not only exhibit high geometric similarity to GT objects but also maintain consistency in style. This demonstrates that our model effectively captures the core features of 3D objects in the process of translating EEG signals into complex visual information, providing evidence that EEG signals can perceive and process various visual information such as color, shape, and texture.

\subsubsection{Quantity}
As shown in Table~\ref{Quantitative_2d} and Table~\ref{table3_3d}, our quantitative analysis results are presented. We primarily used the following metrics to evaluate the performance of the reconstructed 2D and 3D models: 1) Fréchet Inception Distance (FID) ~\cite{FID}, which assesses the difference between the reconstructed 2D images and real images; 2) Structural Similarity Index Measure (SSIM) ~\cite{wang2004image}, which measures the authenticity of the 2D images; 3) Inception Score (IS) ~\cite{Szegedy}, which evaluates the quality and diversity of the reconstructed images; 4) LPIPS ~\cite{zhang2018unreasonable}, which assesses the 3D reconstruction quality of reference views; and 5) Contextual Distance ~\cite{mechrez2018contextual}, which measures the pixel-level similarity between novel view renderings and reference images.

\paragraph{\textbf{Quantitative Analysis of 2D}}Among these metrics, 1), 2), and 3) are quantitative indicators for 2D images. As shown in Table \ref{Quantitative_2d}, some values are missing due to the fact that certain metrics were not reported in the literature for the comparison methods, and the source code was not publicly available. For the Acc metric in the 50-way top-1 task, the accuracy of our proposed method is 21.4\% higher than that of DreamDiffusion ~\cite{BaiECCV24}. In terms of the IS metric, our model also significantly outperforms Brain2Image ~\cite{KavasidisMM17} and NeuroVision ~\cite{Khare}, indicating that the images generated by our model are more realistic and clearer. Compared to these models, our method demonstrates superior performance in both diversity and authenticity of the reconstructed images from EEG signals.

\paragraph{\textbf{Quantitative Analysis of 3D}}Metrics 4) and 5) are qualitative indicators for 3D objects. As shown in Table \ref{table3_3d}, we evaluated the LPIPS and Contextual Distance metrics from different angles during the process of generating style-consistent 3D objects using EEG signals. The Acc metric here refers to the scores obtained using CLIP, where we assess the semantic similarity between the newly generated views and the reference views to measure the results.

\paragraph{\textbf{Ablation Study of 3D}}We conducted three ablation experiments to quantitatively analyze the impact of each component on the visual quality of style-consistent 3D objects. Table \ref{Ablation Study 3d} presents the results of the ablation studies. When all components are utilized, our network achieves the best performance.  

1) ${{\mathcal{L}}_{neuron-style}}$. To enable the EEG reconstruction of style-consistent 3D objects, we introduced neural style loss, which ensures that the 2D images reconstructed from different viewpoints using EEG maintain style consistency with GT. This approach facilitates the generation of style-consistent 3D objects.

2) ${{\mathcal{L}}_{EEG-CLIP}}$. Since the method proposed in this paper aligns EEG signals with novel views from different perspectives, we introduce the alignment between CLIP’s image encoder and EEG signals.  ${{\mathcal{L}}_{EEG-CLIP}}$ aligns the novel view image $g(\theta)$ obtained under the diffusion prior with the EEG signals.

3) ${{\mathcal{L}}_{region}}$. To enable the diffusion model with semantic region-aware capabilities, we keep the inherent ${{\mathcal{L}}_{ldm}}$ property of the LDM unchanged. By integrating latent EEG codes with regional images, we fine-tune LDM, allowing it to reconstruct images with precise positional and spatial representation.

\section{Conclusions}
We propose an EEG-to-3D with style consistency approach, which, for the first time, utilizes EEG signals to generate high-fidelity 3D objects with consistent style. This method employs a two-stage strategy, integrating EEG signals, visual stimulus images, and style loss, to jointly fine-tune LDM and NeRF through collaborative training. The reconstructed 3D objects exhibit a high degree of realism and consistency in both geometry and color.
\paragraph{\textbf{Limitations}}At stage A phase, the semantic accuracy of 2D images reconstructed from EEG signals still has room for improvement, which is expected to further enhance the clarity of the generated 3D objects. Nevertheless, it is noteworthy that the findings of this study have demonstrated that brain signals can perceive and encode various visual features, such as the color, shape, and texture of objects.

\newpage
\appendix
\onecolumn
As shown in Tables \ref{more_eeg2_3d} and \ref{more_imageNet}, we present several key reconstructed images featuring style-consistent 3D objects. Our model utilizes EEG to directly reconstruct these style-consistent 3D objects, accurately restoring the spatial positions of the objects and the spatial relationships between multiple objects within the scene. As illustrated in Table \ref{fmri_3d}, the method proposed in this paper has been validated on the fMRI dataset, demonstrating the ability to reconstruct style-consistent 3D objects using fMRI data.

\begin{table*}[hbt]
    \small
    \centering
    % table caption is above the table
    % For LaTeX tables use
    \begin{tabular}{ccccc}
        \hline
        Visual Stimulus Images     &with ${{\mathcal{L}}_{neuron-style}}$  & 3D Views (${20}^{o}$)   &${45}^{o}$  &${70}^{o}$ \\
        \hline
        \makecell[c]{\\
        \begin{minipage}[b]{0.2\columnwidth}
            \centering
            {\includegraphics[width=1\textwidth]{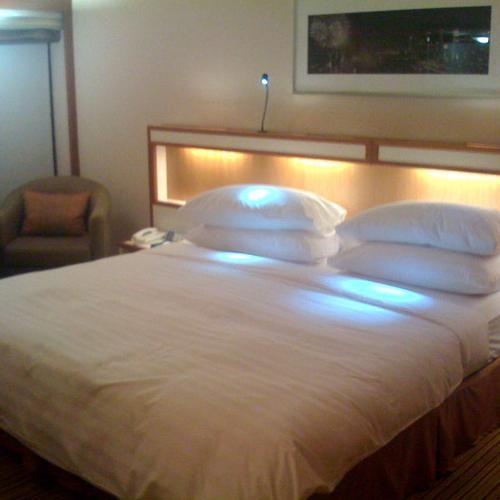}}
        \end{minipage}
        }
        &\makecell[c]{\\
        \begin{minipage}[b]{0.2\columnwidth}
            \centering
            {\includegraphics[width=1\textwidth]{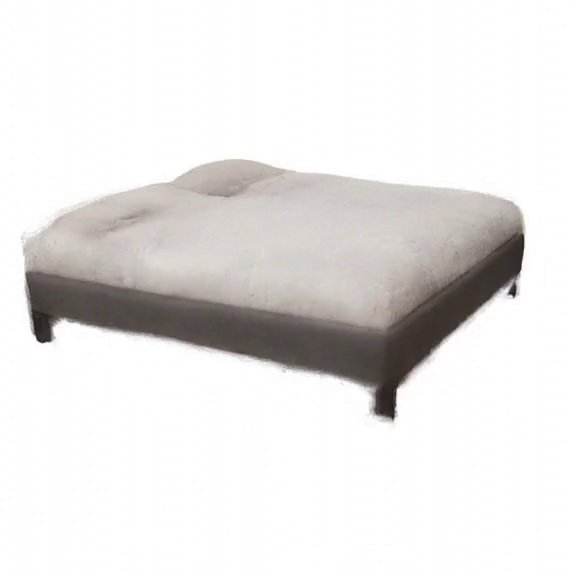}}
        \end{minipage}
        }  
        &\makecell[c]{\\
        \begin{minipage}[b]{0.2\columnwidth}
            \centering
            {\includegraphics[width=1\textwidth]{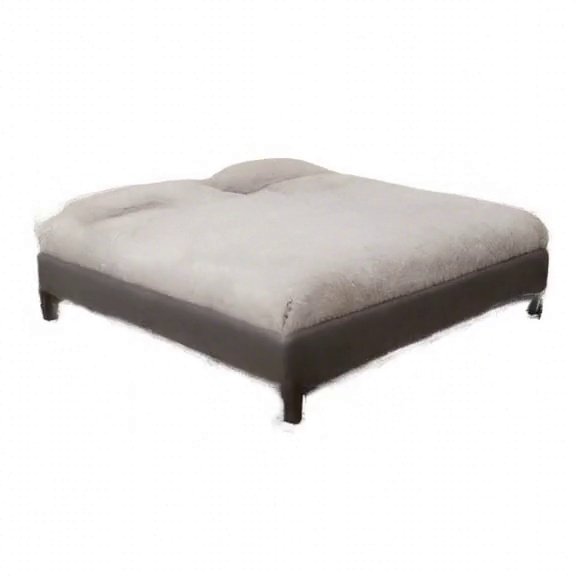}}
        \end{minipage}
        } 
        & \makecell[c]{\\
        \begin{minipage}[b]{0.2\columnwidth}
            \centering
            {\includegraphics[width=1\textwidth]{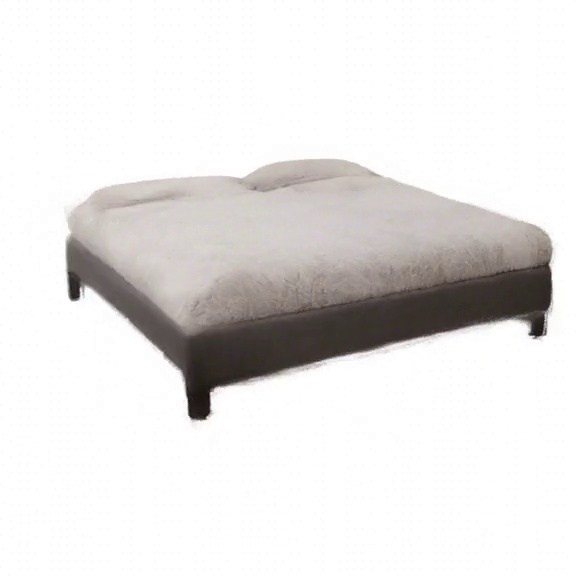}}
        \end{minipage}
        }
        & \makecell[c]{\\
        \begin{minipage}[b]{0.2\columnwidth}
            \centering
            {\includegraphics[width=1\textwidth]{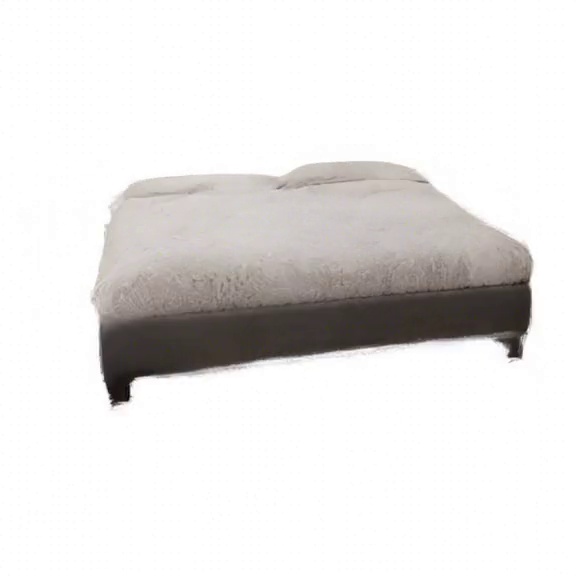}}
        \end{minipage}
        }
        \\
        \makecell[c]{\\
        \begin{minipage}[b]{0.2\columnwidth}
            \centering
            {\includegraphics[width=0.7\textwidth]{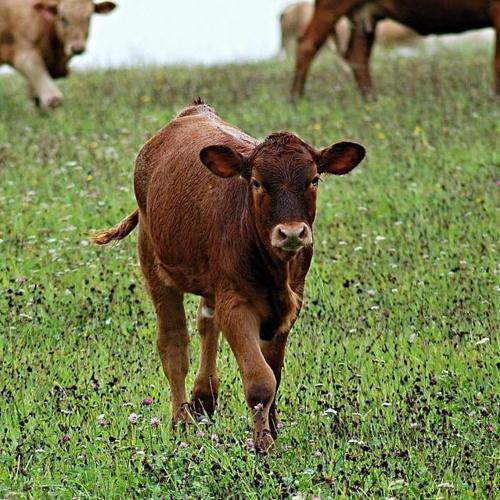}}
        \end{minipage}
        }
        &\makecell[c]{\\
        \begin{minipage}[b]{0.2\columnwidth}
            \centering
            {\includegraphics[width=0.7\textwidth]{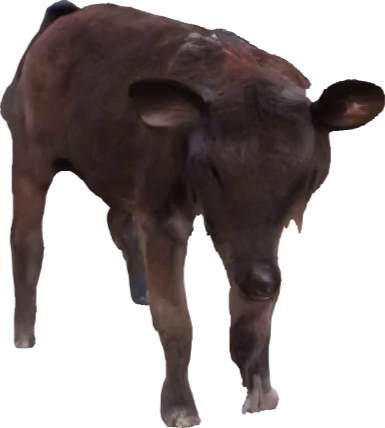}}
        \end{minipage}
        }  
        &\makecell[c]{\\
        \begin{minipage}[b]{0.2\columnwidth}
            \centering
            {\includegraphics[width=0.7\textwidth]{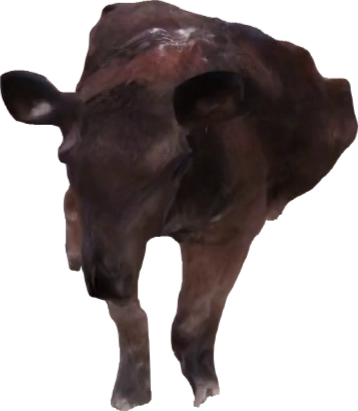}}
        \end{minipage}
        } 
        & \makecell[c]{\\
        \begin{minipage}[b]{0.2\columnwidth}
            \centering
            {\includegraphics[width=0.7\textwidth]{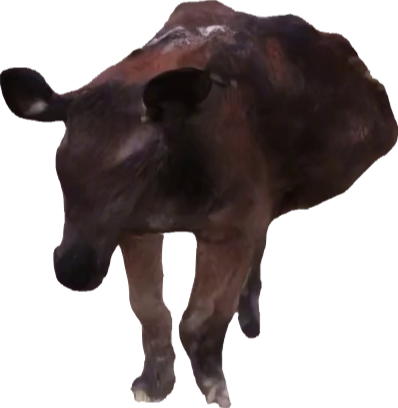}}
        \end{minipage}
        }
        & \makecell[c]{\\
        \begin{minipage}[b]{0.2\columnwidth}
            \centering
            {\includegraphics[width=0.7\textwidth]{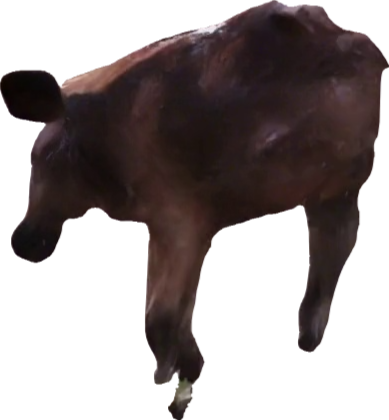}}
        \end{minipage}
        }
        \\
        \makecell[c]{\\
        \begin{minipage}[b]{0.2\columnwidth}
            \centering
            {\includegraphics[width=1\textwidth]{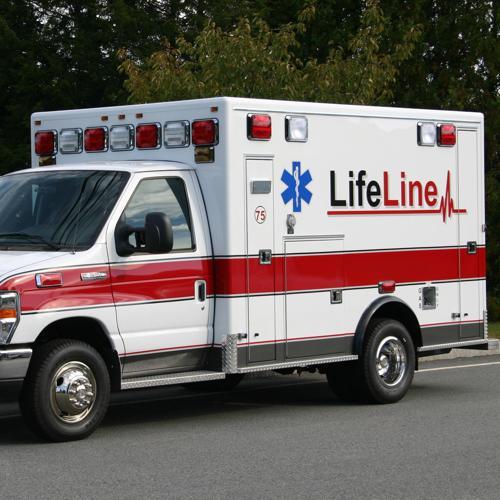}}
        \end{minipage}
        }
        &\makecell[c]{\\
        \begin{minipage}[b]{0.2\columnwidth}
            \centering
            {\includegraphics[width=1\textwidth]{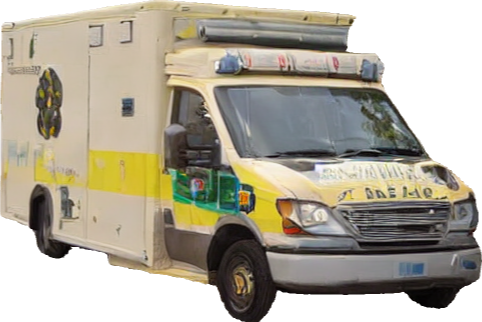}}
        \end{minipage}
        }  
        &\makecell[c]{\\
        \begin{minipage}[b]{0.2\columnwidth}
            \centering
            {\includegraphics[width=1\textwidth]{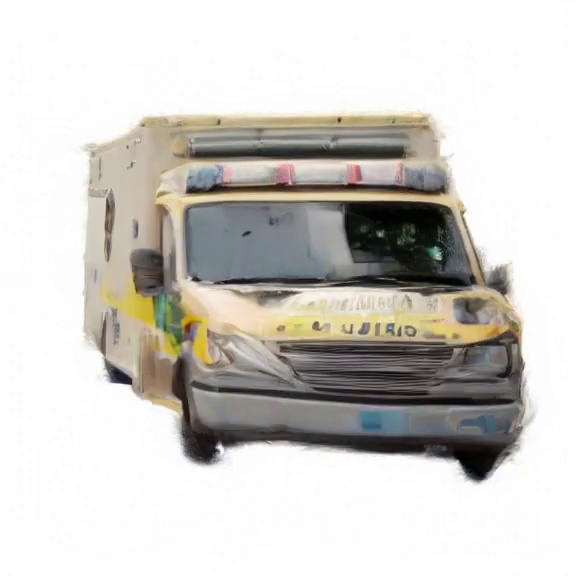}}
        \end{minipage}
        } 
        & \makecell[c]{\\
        \begin{minipage}[b]{0.2\columnwidth}
            \centering
            {\includegraphics[width=1\textwidth]{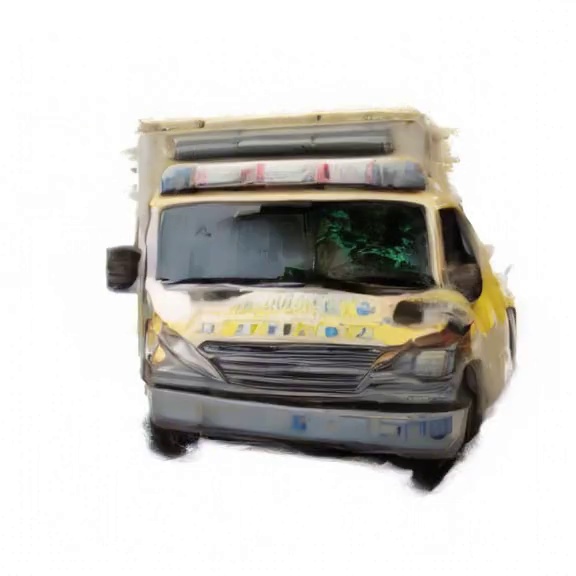}}
        \end{minipage}
        }
        & \makecell[c]{\\
        \begin{minipage}[b]{0.2\columnwidth}
            \centering
            {\includegraphics[width=1\textwidth]{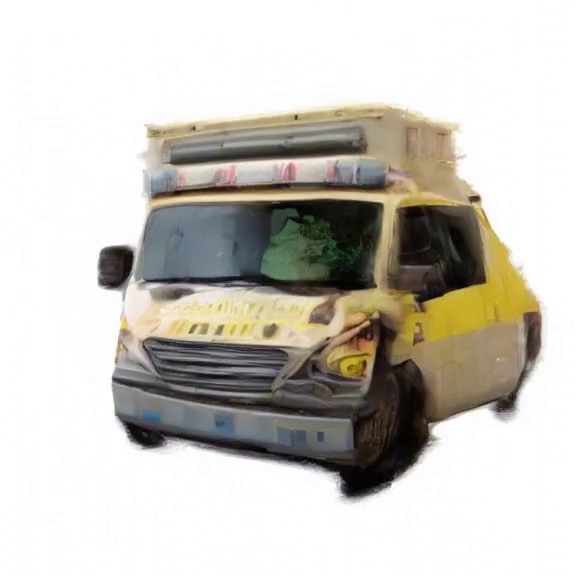}}
        \end{minipage}
        }
        \\
        \hline
    \end{tabular}
    \caption{Additional typical 3D objects generated by EEG-driven models based on the Things-EEG2 dataset ~\cite{gifford2022large}.}
    \label{more_eeg2_3d}
\end{table*}

\begin{table*}[hbt]
    \small
    \centering
    % table caption is above the table
    % For LaTeX tables use
    \begin{tabular}{ccccc}
        \hline
        Visual Stimulus Images     &with ${{\mathcal{L}}_{neuron-style}}$  & 3D Views (${20}^{o}$)   &${45}^{o}$  &${70}^{o}$ \\
        \hline
        \makecell[c]{\\
        \begin{minipage}[b]{0.2\columnwidth}
            \centering
            {\includegraphics[width=1\textwidth]{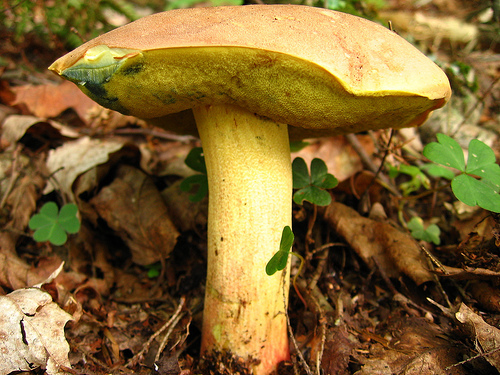}}
        \end{minipage}
        }
        &\makecell[c]{\\
        \begin{minipage}[b]{0.2\columnwidth}
            \centering
            {\includegraphics[width=0.7\textwidth]{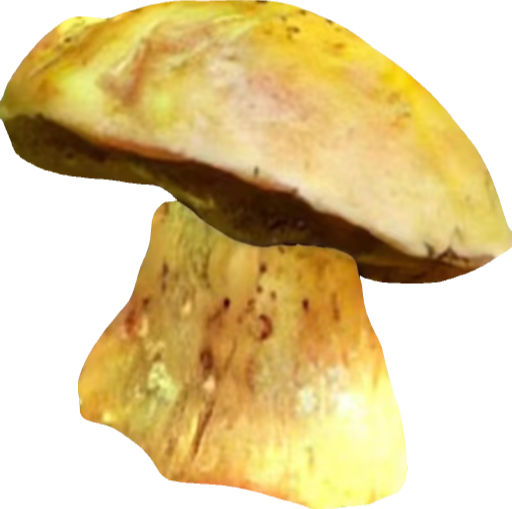}}
        \end{minipage}
        }  
        &\makecell[c]{\\
        \begin{minipage}[b]{0.2\columnwidth}
            \centering
            {\includegraphics[width=0.7\textwidth]{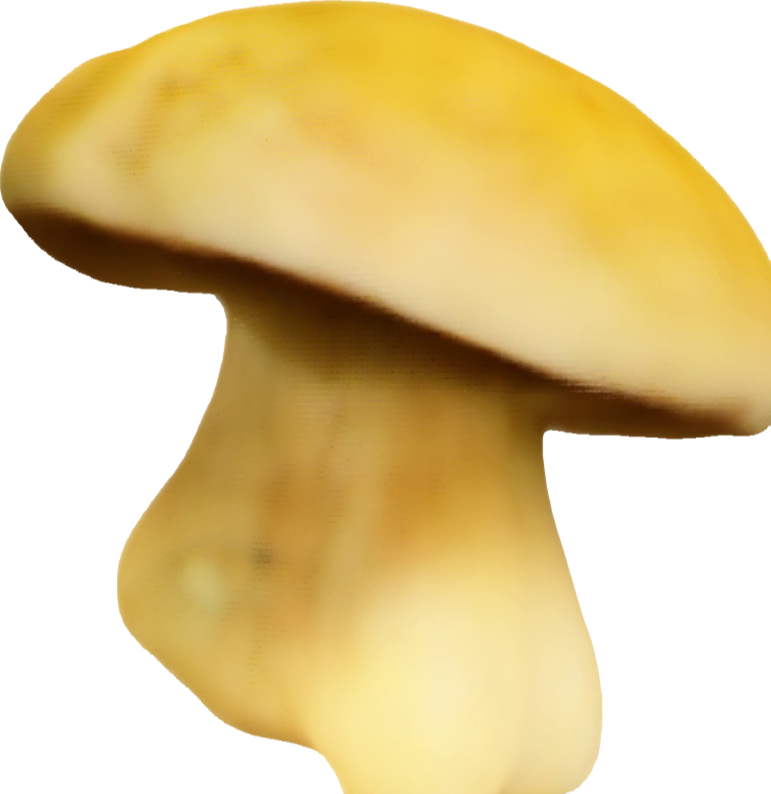}}
        \end{minipage}
        } 
        & \makecell[c]{\\
        \begin{minipage}[b]{0.2\columnwidth}
            \centering
            {\includegraphics[width=0.7\textwidth]{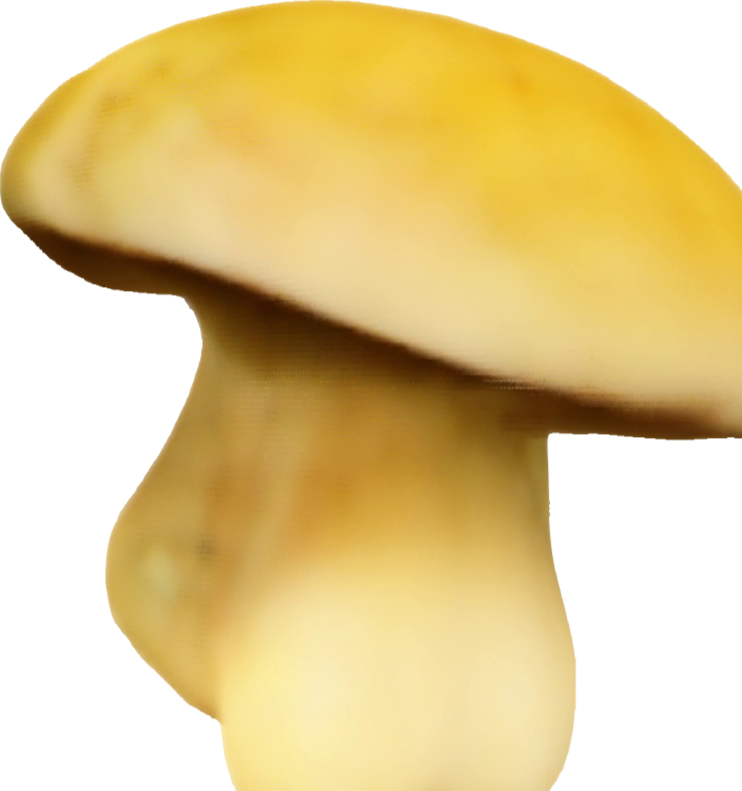}}
        \end{minipage}
        }
        & \makecell[c]{\\
        \begin{minipage}[b]{0.2\columnwidth}
            \centering
            {\includegraphics[width=0.7\textwidth]{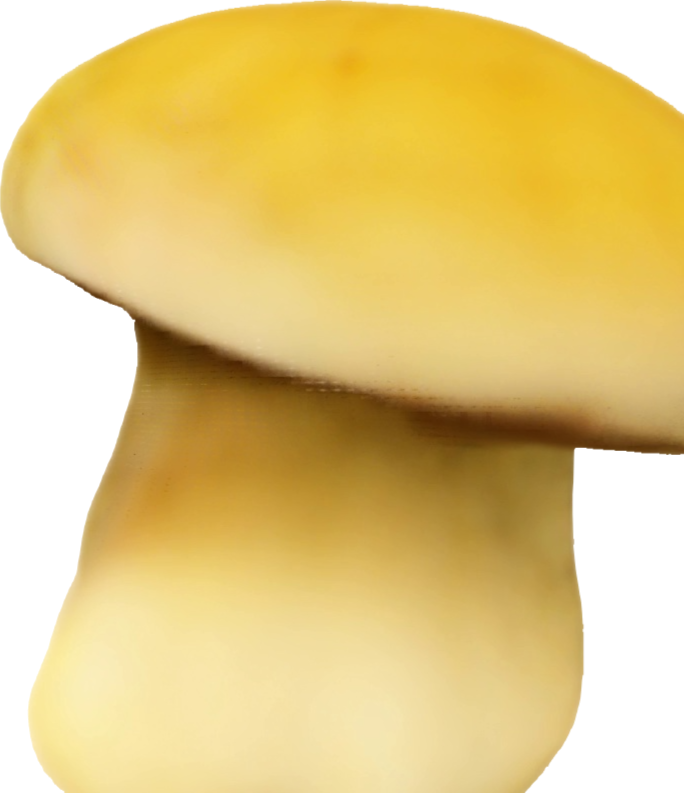}}
        \end{minipage}
        }        
        \\
        \makecell[c]{\\
        \begin{minipage}[b]{0.2\columnwidth}
            \centering
            {\includegraphics[width=1\textwidth]{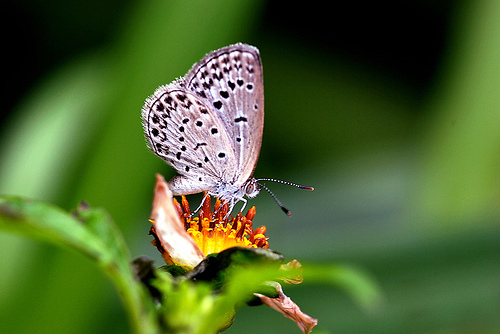}}
        \end{minipage}
        }
        &\makecell[c]{\\
        \begin{minipage}[b]{0.2\columnwidth}
            \centering
            {\includegraphics[width=0.7\textwidth]{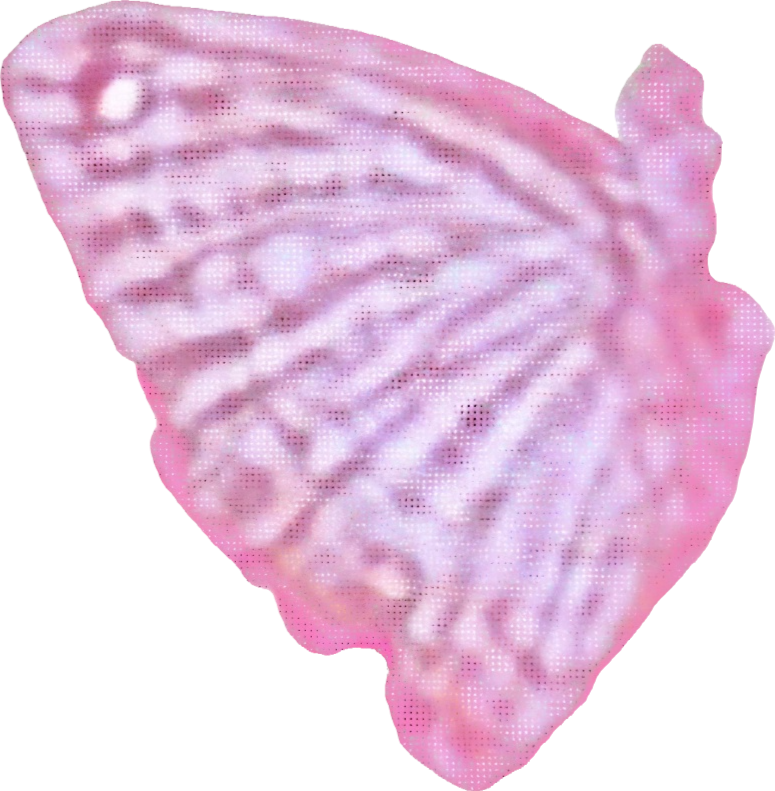}}
        \end{minipage}
        }  
        &\makecell[c]{\\
        \begin{minipage}[b]{0.2\columnwidth}
            \centering
            {\includegraphics[width=0.7\textwidth]{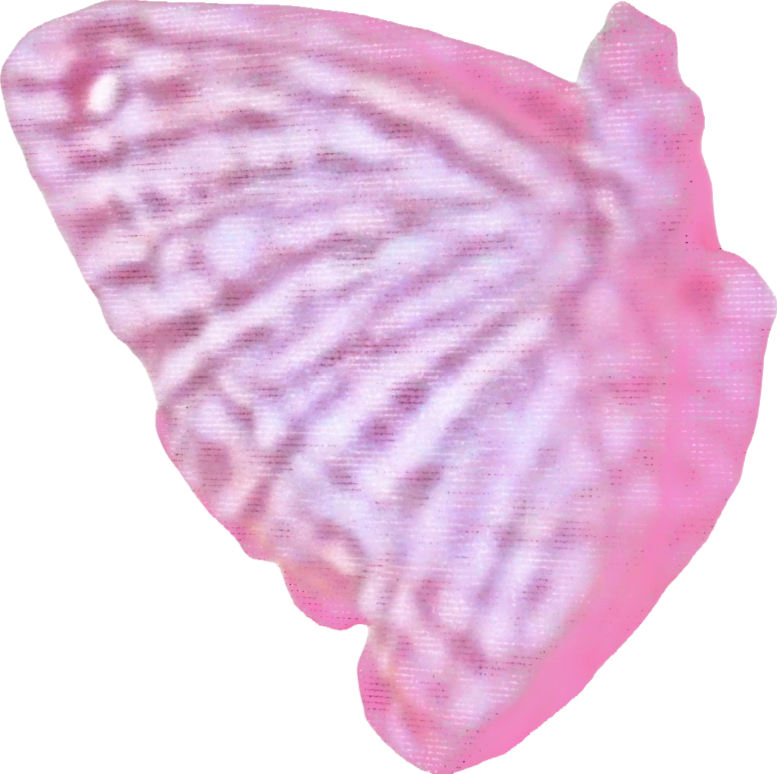}}
        \end{minipage}
        } 
        & \makecell[c]{\\
        \begin{minipage}[b]{0.2\columnwidth}
            \centering
            {\includegraphics[width=0.7\textwidth]{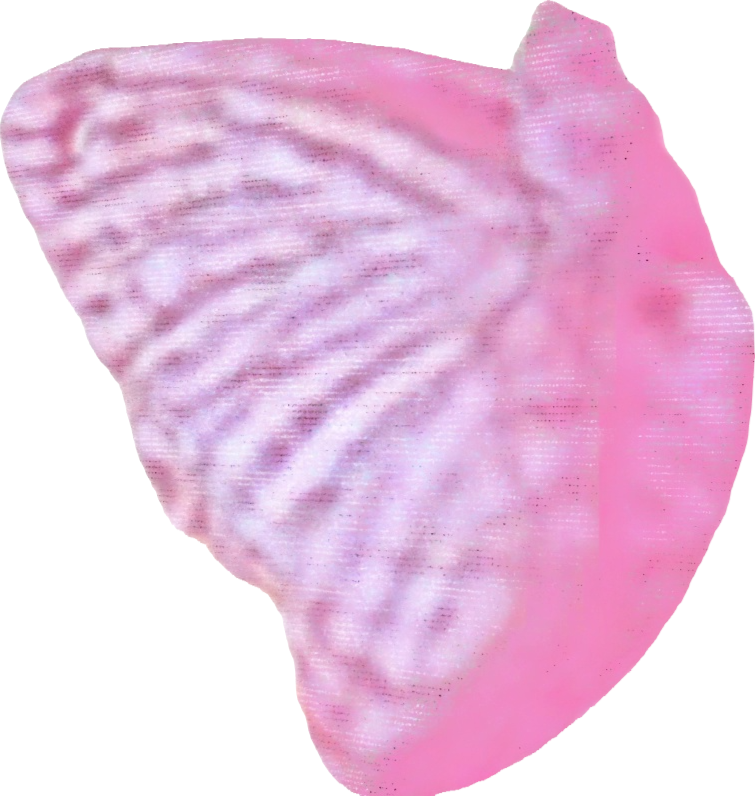}}
        \end{minipage}
        }
        & \makecell[c]{\\
        \begin{minipage}[b]{0.2\columnwidth}
            \centering
            {\includegraphics[width=0.7\textwidth]{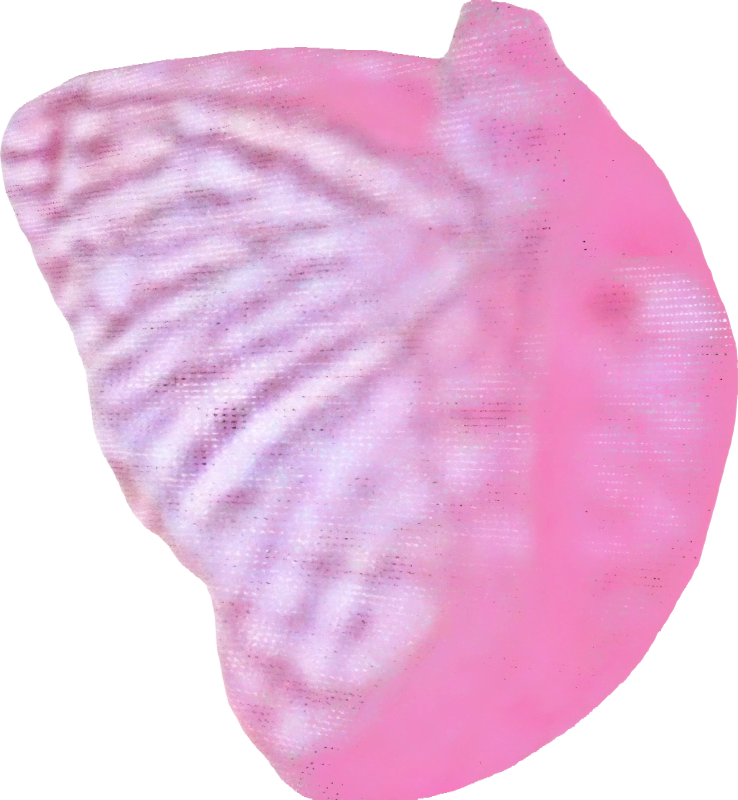}}
        \end{minipage}
        }
        \\      
        \makecell[c]{\\
        \begin{minipage}[b]{0.2\columnwidth}
            \centering
            {\includegraphics[width=1\textwidth]{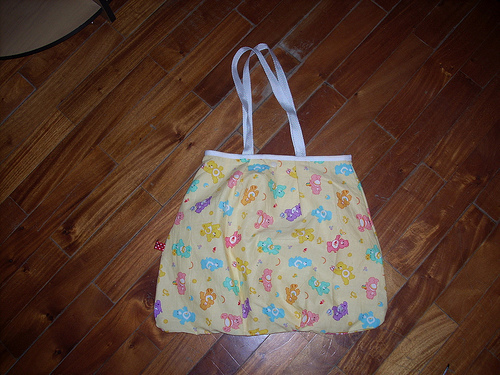}}
        \end{minipage}
        }
        &\makecell[c]{\\
        \begin{minipage}[b]{0.2\columnwidth}
            \centering
            {\includegraphics[width=0.7\textwidth]{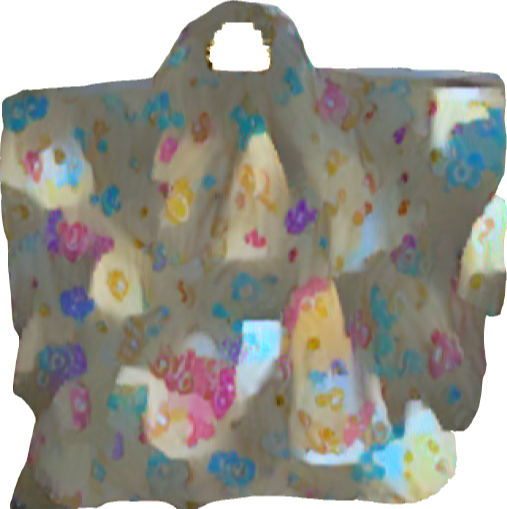}}
        \end{minipage}
        }  
        &\makecell[c]{\\
        \begin{minipage}[b]{0.2\columnwidth}
            \centering
            {\includegraphics[width=0.7\textwidth]{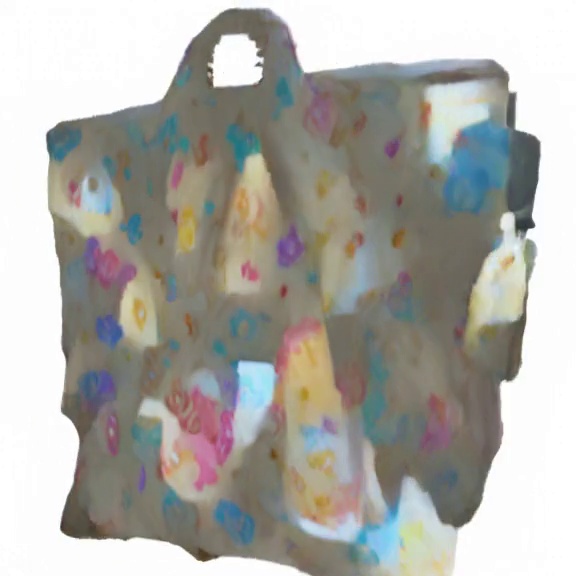}}
        \end{minipage}
        } 
        & \makecell[c]{\\
        \begin{minipage}[b]{0.2\columnwidth}
            \centering
            {\includegraphics[width=0.7\textwidth]{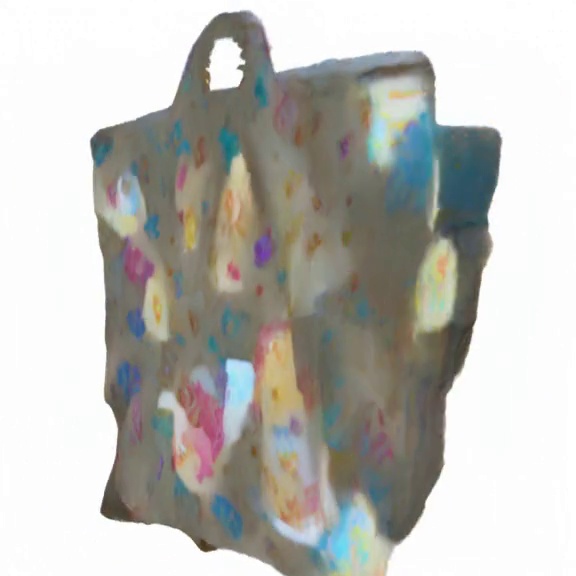}}
        \end{minipage}
        }
        & \makecell[c]{\\
        \begin{minipage}[b]{0.2\columnwidth}
            \centering
            {\includegraphics[width=0.7\textwidth]{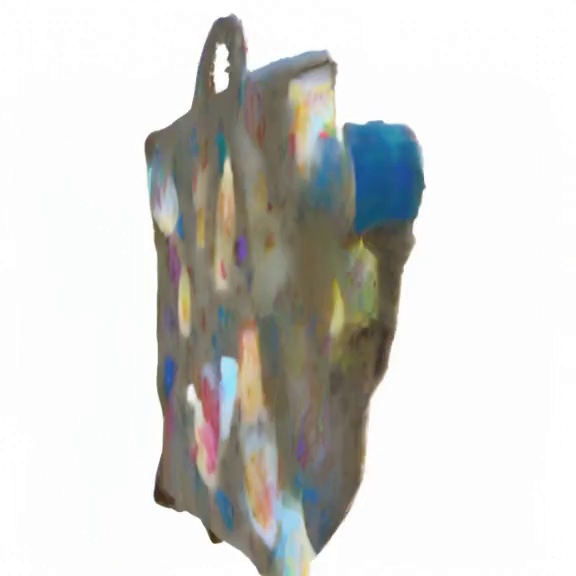}}
        \end{minipage}
        }
        \\
        \hline
    \end{tabular}
    \caption{Additional typical 3D objects generated by EEG-driven models based on the EEG-ImageNet dataset ~\cite{spampinato2017deep}.}
    \label{more_imageNet}
\end{table*}

\begin{table*}[hbt]
    \small
    \centering
    % table caption is above the table
    % For LaTeX tables use
    \begin{tabular}{ccccc}
        \hline
        Visual Stimulus Images     &with ${{\mathcal{L}}_{neuron-style}}$  & 3D Views (${20}^{o}$)   &${45}^{o}$  &${70}^{o}$ \\
        \hline
        \makecell[c]{\\
        \begin{minipage}[b]{0.2\columnwidth}
            \centering
            {\includegraphics[width=1\textwidth]{red_car.JPEG}}
        \end{minipage}
        }
        &\makecell[c]{\\
        \begin{minipage}[b]{0.2\columnwidth}
            \centering
            {\includegraphics[width=0.5\textwidth]{red_car_color.png}}
        \end{minipage}
        }  
        &\makecell[c]{\\
        \begin{minipage}[b]{0.2\columnwidth}
            \centering
            {\includegraphics[width=0.7\textwidth]{car_red_1.png}}
        \end{minipage}
        } 
        & \makecell[c]{\\
        \begin{minipage}[b]{0.2\columnwidth}
            \centering
            {\includegraphics[width=0.7\textwidth]{car_red_2.png}}
        \end{minipage}
        }
        & \makecell[c]{\\
        \begin{minipage}[b]{0.2\columnwidth}
            \centering
            {\includegraphics[width=0.7\textwidth]{car_red_3.png}}
        \end{minipage}
        }        
        \\        
        \makecell[c]{\\
        \begin{minipage}[b]{0.2\columnwidth}
            \centering
            {\includegraphics[width=0.7\textwidth]{xiyiji.JPEG}}
        \end{minipage}
        }
        &\makecell[c]{\\
        \begin{minipage}[b]{0.2\columnwidth}
            \centering
            {\includegraphics[width=0.7\textwidth]{xiyiji_color.png}}
        \end{minipage}
        }  
        &\makecell[c]{\\
        \begin{minipage}[b]{0.2\columnwidth}
            \centering
            {\includegraphics[width=0.7\textwidth]{xiyiji_1.png}}
        \end{minipage}
        } 
        & \makecell[c]{\\
        \begin{minipage}[b]{0.2\columnwidth}
            \centering
            {\includegraphics[width=0.7\textwidth]{xiyiji_2.png}}
        \end{minipage}
        }
        & \makecell[c]{\\
        \begin{minipage}[b]{0.2\columnwidth}
            \centering
            {\includegraphics[width=0.7\textwidth]{xiyiji_3.png}}
        \end{minipage}
        }
        \\    
        \makecell[c]{\\
        \begin{minipage}[b]{0.2\columnwidth}
            \centering
            {\includegraphics[width=1\textwidth]{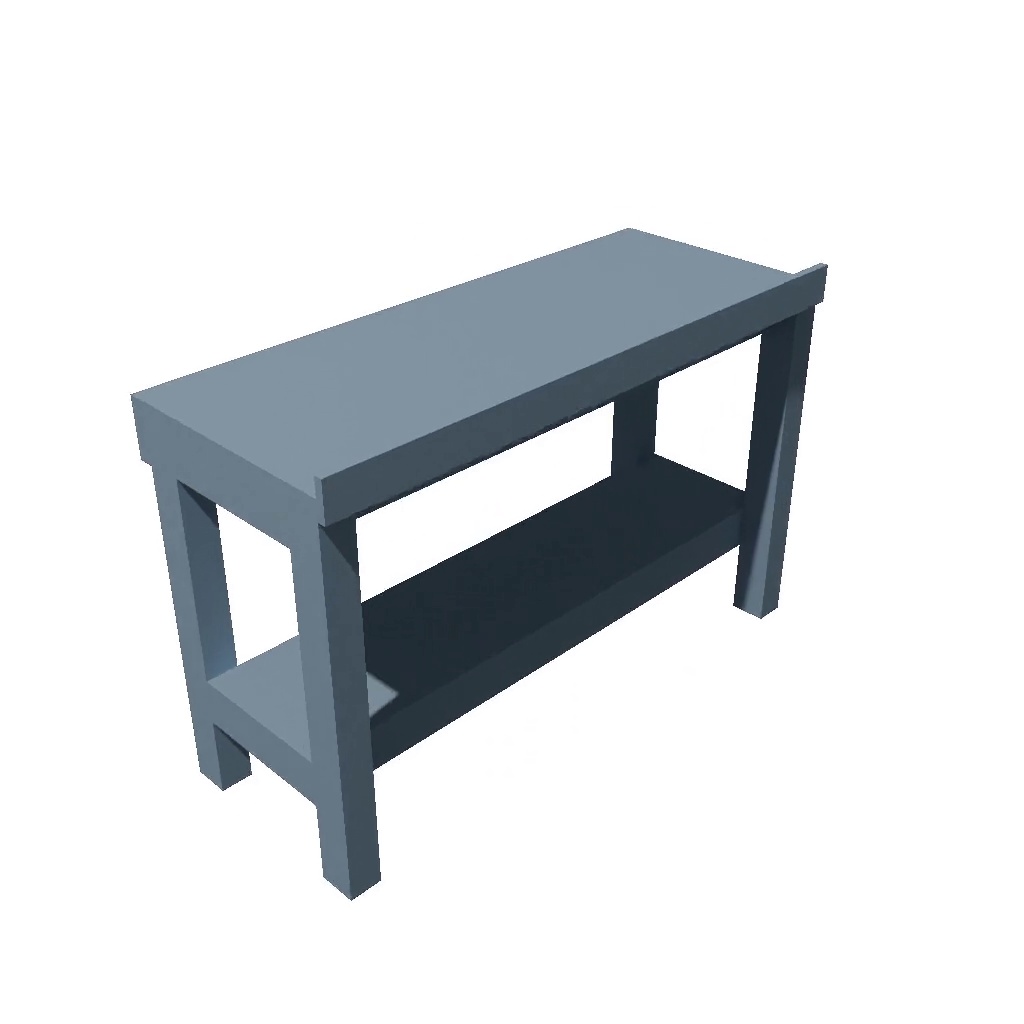}}
        \end{minipage}
        }
        &\makecell[c]{\\
        \begin{minipage}[b]{0.2\columnwidth}
            \centering
            {\includegraphics[width=0.7\textwidth]{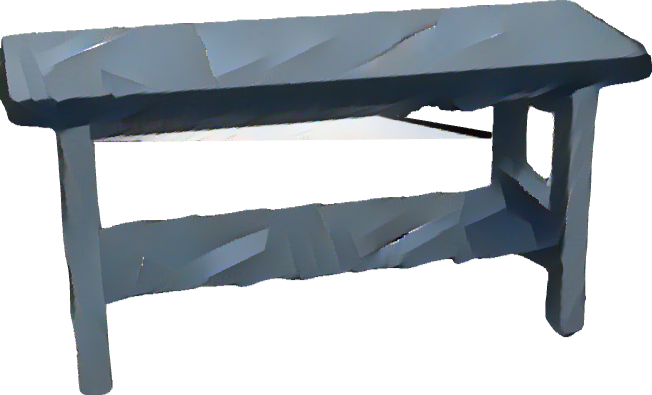}}
        \end{minipage}
        }  
        &\makecell[c]{\\
        \begin{minipage}[b]{0.2\columnwidth}
            \centering
            {\includegraphics[width=0.7\textwidth]{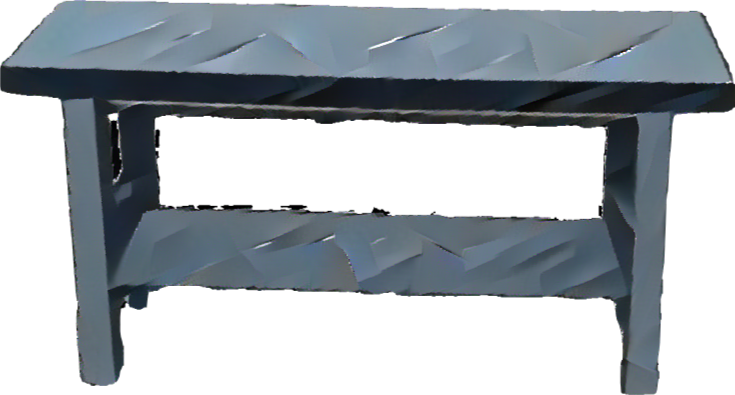}}
        \end{minipage}
        } 
        & \makecell[c]{\\
        \begin{minipage}[b]{0.2\columnwidth}
            \centering
            {\includegraphics[width=0.7\textwidth]{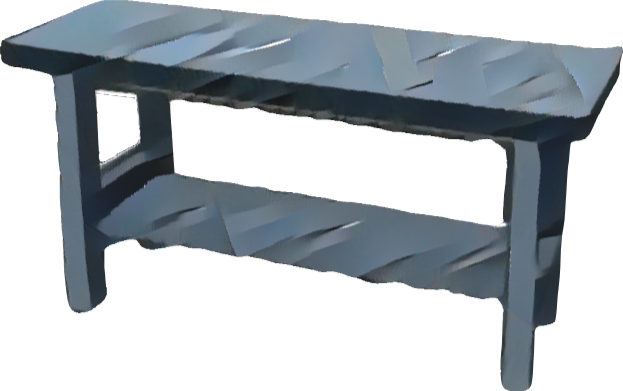}}
        \end{minipage}
        }
        & \makecell[c]{\\
        \begin{minipage}[b]{0.2\columnwidth}
            \centering
            {\includegraphics[width=0.7\textwidth]{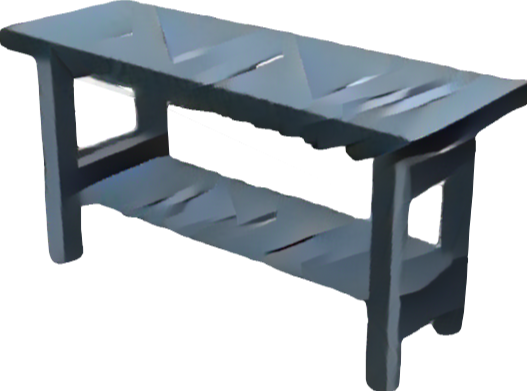}}
        \end{minipage}
        }
        \\
        \hline
    \end{tabular}
    \caption{Additional typical 3D objects generated by ours models based on the fmri dataset ~\cite{gao2023mind}.}
    \label{fmri_3d}
\end{table*}

\begin{figure*}
    \centering
    \includegraphics[width=0.7\linewidth]{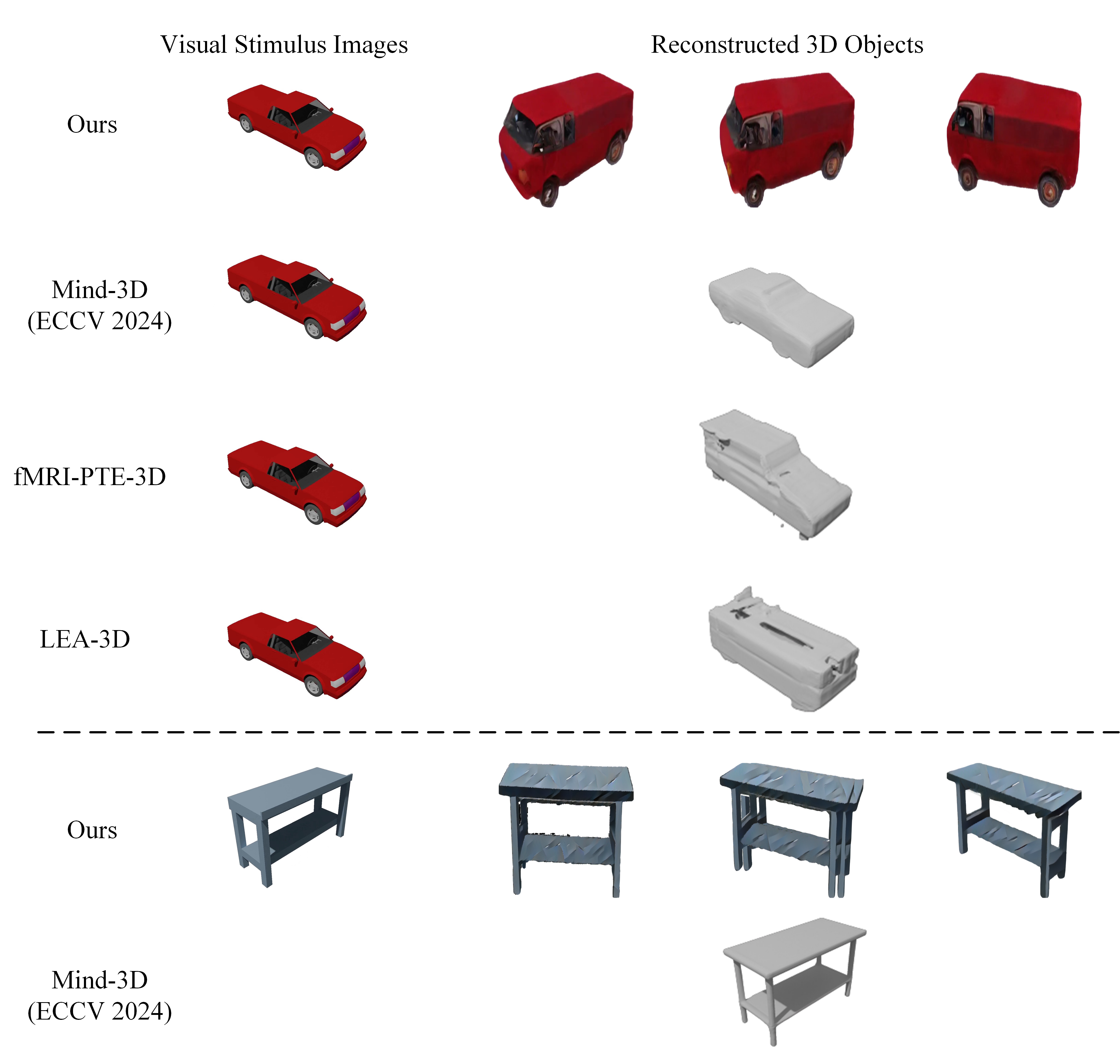}
    \caption{Our model reconstructs 3D objects using fMRI.}
    \label{fmri}
\end{figure*}

\bibliographystyle{unsrt}

\end{document}